\documentclass[journal]{IEEEtran}
\usepackage{balance}
\usepackage{amsfonts}
\usepackage{float}
\usepackage[pdftex]{graphicx}
\usepackage{subfigure}
\usepackage{epstopdf}
\usepackage[colorlinks,linkcolor=red]{hyperref}
\ifCLASSOPTIONcompsoc
\usepackage{booktabs}
  \usepackage[nocompress]{cite}
\else
  \usepackage{cite}
\fi
\usepackage{algorithm}
\usepackage{algorithmic}
\usepackage{amsmath}
\usepackage{makecell}
\usepackage{multirow}
\usepackage{color}

\begin{document}

\title{Bounding Boxes Are All We Need: \\Street View Image Classification via \\Context Encoding of Detected Buildings}
\newcommand{\w}{\textcolor{black}}
\newcommand{\s}{\textcolor{red}}
\author{Kun Zhao,~
	Yongkun Liu,~          
	Siyuan Hao,~
	Shaoxing Lu,~
	Hongbin Liu,~
	Lijian Zhou % <-this % stops a space
	\thanks{\textit{Corresponding author: Lijian Zhou.}}
	%\thanks{Corresponding author: Yuanxin Ye.}
	\thanks{Kun Zhao, Yongkun Liu, Siyuan Hao, Shaoxing Lu and Lijian Zhou were with the School of Information and Control Engineering, Qingdao University of Technology, Qingdao 266520, China. 
		E-mail: sterling1982@163.com, YongkunLiu.mail@gmail.com, lemonbananan@163.com, 2445252341@qq.com, zhoulijian@qut.edu.cn}% <-this % stops a space
	\thanks{Hongbin Liu was with the BIM Research Center, Qingdao Research Institute of Urban and Rural Construction,  Qingdao 266033, China. 
		E-mail: binbin\_sky@163.com}% <-this % stops a space
}
% The paper headers
\markboth{Kun Zhao \MakeLowercase{\textit{et al.}}: Bounding Boxes Are All We Need: Street View Image Classification via Context Encoding of Detected Buildings}{}
\maketitle

\begin{abstract}
Street view images classification aiming at urban land use analysis is difficult because the class labels (e.g., commercial area), are concepts with higher abstract level compared to the ones of general visual tasks (e.g., persons and cars). Therefore, classification models using only visual features often fail to achieve satisfactory performance. In this paper, a novel approach based on a “Detector-Encoder-Classifier” framework is proposed. Instead of using visual features of the whole image directly as common image-level models based on convolutional neural networks (CNNs) do, the proposed framework firstly obtains the bounding boxes of buildings in street view images from a detector. Their contextual information such as the co-occurrence patterns of building classes and their layout are then encoded into metadata by the proposed algorithm “CODING” (Context encOding of Detected buildINGs). Finally, these bounding box metadata are classified by a recurrent neural network (RNN). In addition, we made a dual-labeled dataset named “BEAUTY” (Building dEtection And Urban funcTional-zone portraYing) of 19,070 street view images and 38,857 buildings based on the existing BIC\_GSV~\cite{kang2018building}. The dataset can be used not only for street view image classification, but also for multi-class building detection. Experiments on “BEAUTY” show that the proposed approach achieves a 12.65\% performance improvement on macro-precision and 12\% on macro-recall over image-level CNN based models. Our code and dataset are available at {\small\url{https://github.com/kyle-one/Context-Encoding-of-Detected-Buildings/}}
\end{abstract}


\IEEEpeerreviewmaketitle

\section{Introduction}

\begin{figure}[!htb]
	\centering
	\includegraphics[width=0.9\linewidth]{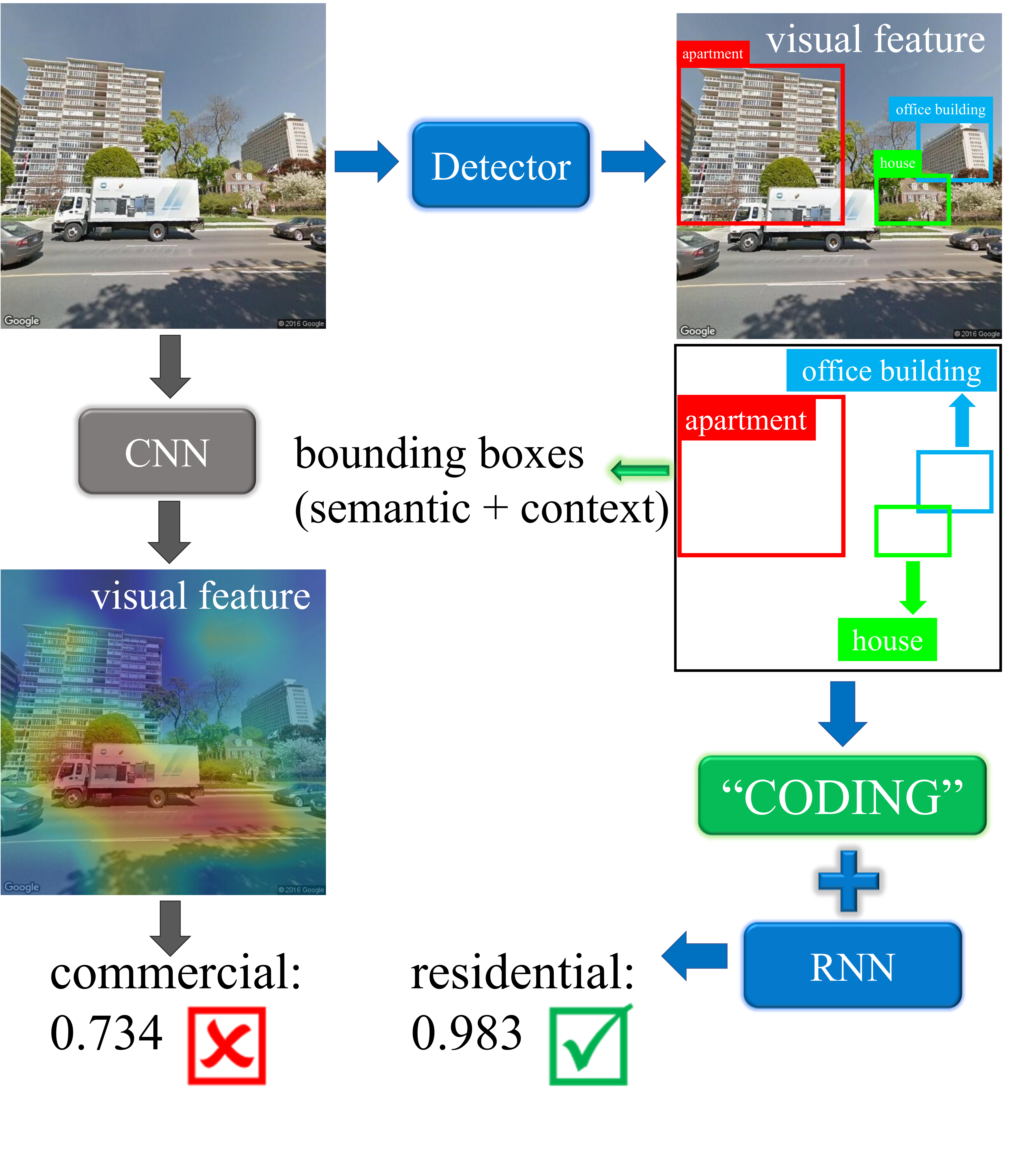}
	\caption{The main idea of this paper. Left: A common image-level CNN based model. Because of the high-level abstractness of the class labels, the visual features learned by CNN are inaccurate (see the heatmap of the last feature layer), which led to wrong predictions. Right: By using a trained detector, the proposed approach obtained bounding boxes of buildings in the input image, which contain semantic labels and their context information such as the co-occurrence patterns and layout. Correct results were obtained by encoding these information using the proposed “CODING” algorithm and an RNN.}
	\label{fig:Motivations}
\end{figure}

\begin{figure}[t]
	\centering
	\subfigure[A religious area locates round 51.022962, -114.08326.]
	{\begin{minipage}[b]{0.38\textwidth}
		\includegraphics[width=\linewidth]{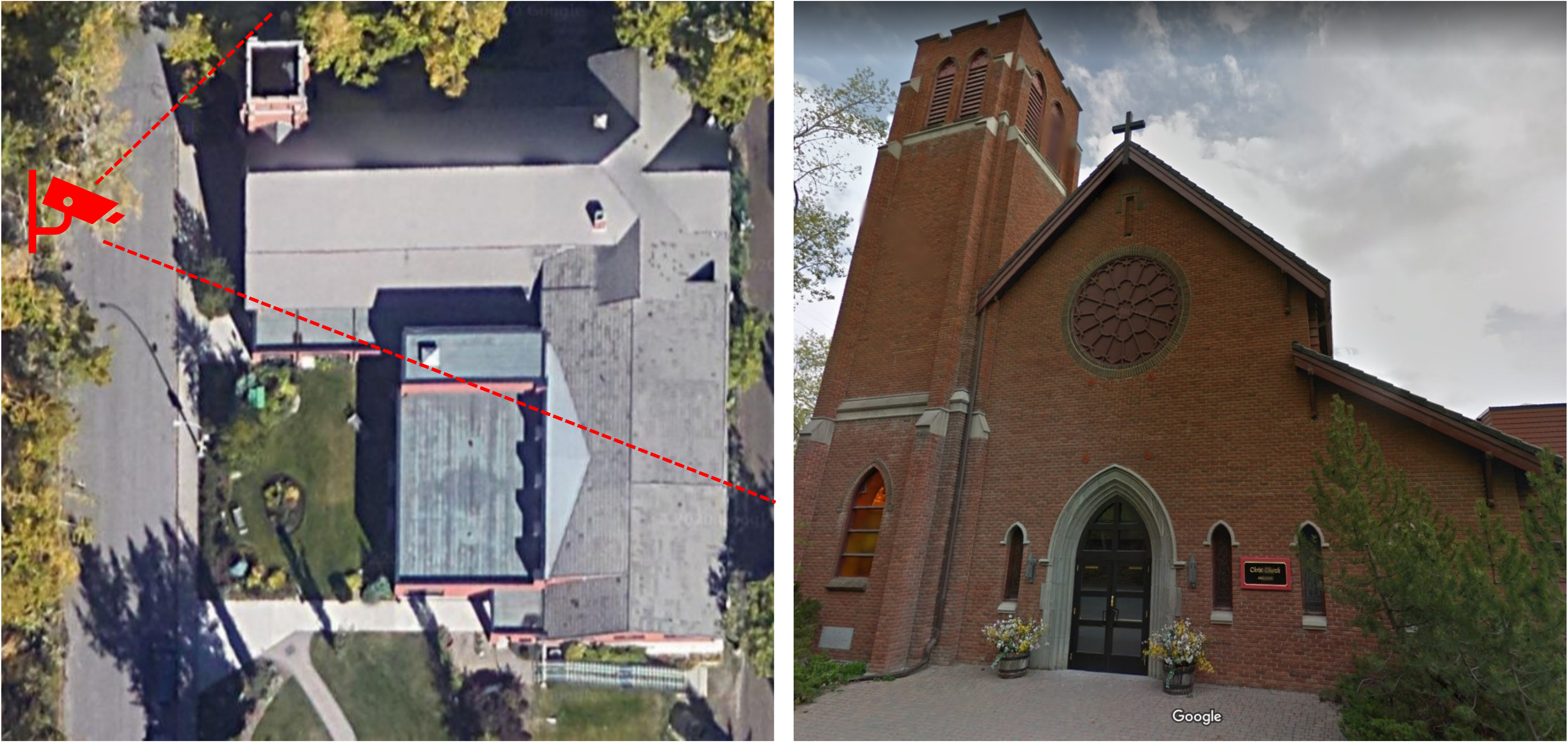}        
	\end{minipage}
    }
    \subfigure[A residential area locates round 51.029009, -114.07783.]
	{\begin{minipage}[b]{0.38\textwidth}
		\includegraphics[width=\linewidth]{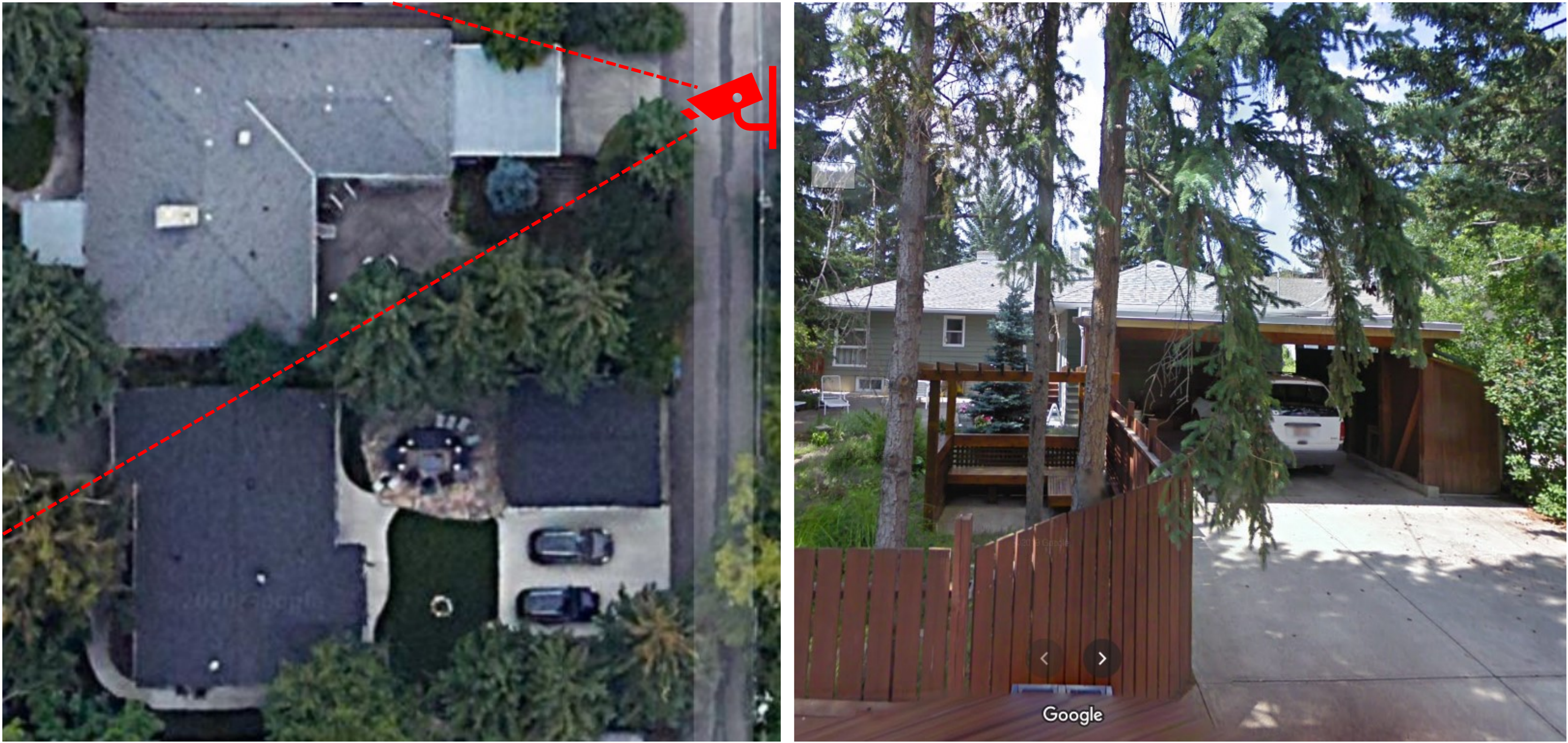}       
	\end{minipage}
    }
	\caption{Areas of different land use with similar looking from overhead view but distinguishable looking from street view.}
	\label{fig:OverStreet}
\end{figure}

\IEEEPARstart{U}{rban} land use records how people use the land with social-economic purposes, such as residential, commercial, and recreational purposes~\cite{cao2018integrating}. Land use classification using satellite images have been extensively studied in remote sensing community. With the rise of geo-data commercial services (e.g., Google maps) and crowdsourced projects (e.g., OpenStreetMap)~\cite{vargas2020openstreetmap}, urban spatial data of different modalities are used~\cite{lefevre2017toward}. As their representative, Google street view (GSV)~\cite{anguelov2010google} provides abundant street-level details which have been increasingly used in urban land use classification. Street view images are accurately geo-located, updated regularly, easy and free to access. Moreover, they contain richer visual information which makes it easier to be distinguished (see Fig.~\ref{fig:OverStreet}). Therefore, visual models that perform well in common computer vision tasks, such like CNNs have recently been widely used to extract visual features of street view images for urban land use and urban functional zone analysis~\cite{cao2018integrating,srivastava2020fine,kang2018building,zhu2019fine,ilic2019deep,hoffmann2019model,srivastava2019understanding,lobry2020rsvqa,liu2020deepsat}. However, the performance so far has been less than satisfactory partly due to the high-level abstractness of urban land use labels, which makes it hard to represent the concepts directly using visual features. In addition, street view images contain many of the same visual elements (e.g., sky and ground) which interfere with distinguishing different usages of land. When using the whole images directly, the most distinguishable visual elements are underutilized. 

\subsection{Motivation}
\label{subsec:Motivation}

We consider street view image classification as a fine-grained outdoor scene analysis problem. The keys of this task are, firstly, acquiring the most significant objects in street view images aiming at land use, and then, effective modeling of their contextual information. Based on the above viewpoints, a “Detector-Encoder-Classifier” framework is proposed to replace the common CNN based architecture, as shown in Fig~\ref{fig:Motivations}.

\begin{figure}[t]
	\centering
	\subfigure[A common object-level label system without subclass of “building”.]
	{\begin{minipage}[t]{0.48\textwidth}
			\includegraphics[width=\linewidth]{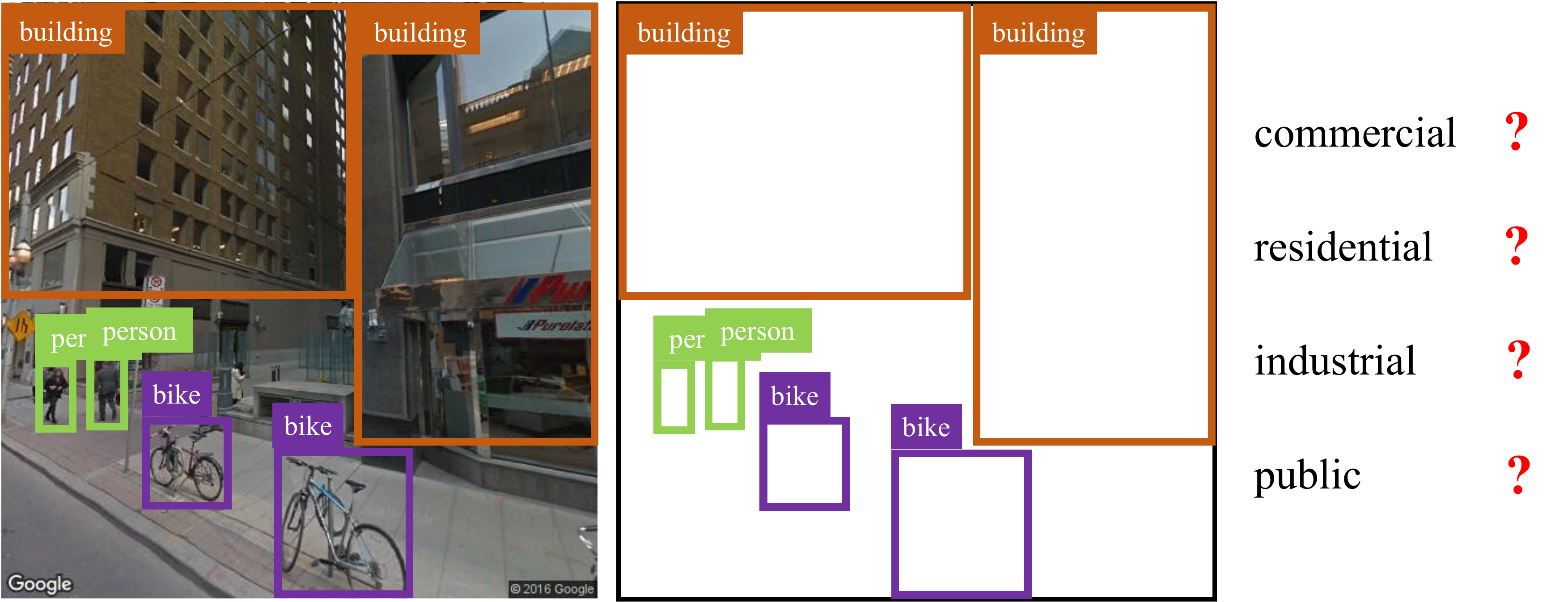} 
			\label{fig:CommonLabel}       
		\end{minipage}
	}
	\subfigure[A land-use oriented label system which takes subclasses of “building” as annotated objects.]
	{\begin{minipage}[t]{0.48\textwidth}
			\includegraphics[width=\linewidth]{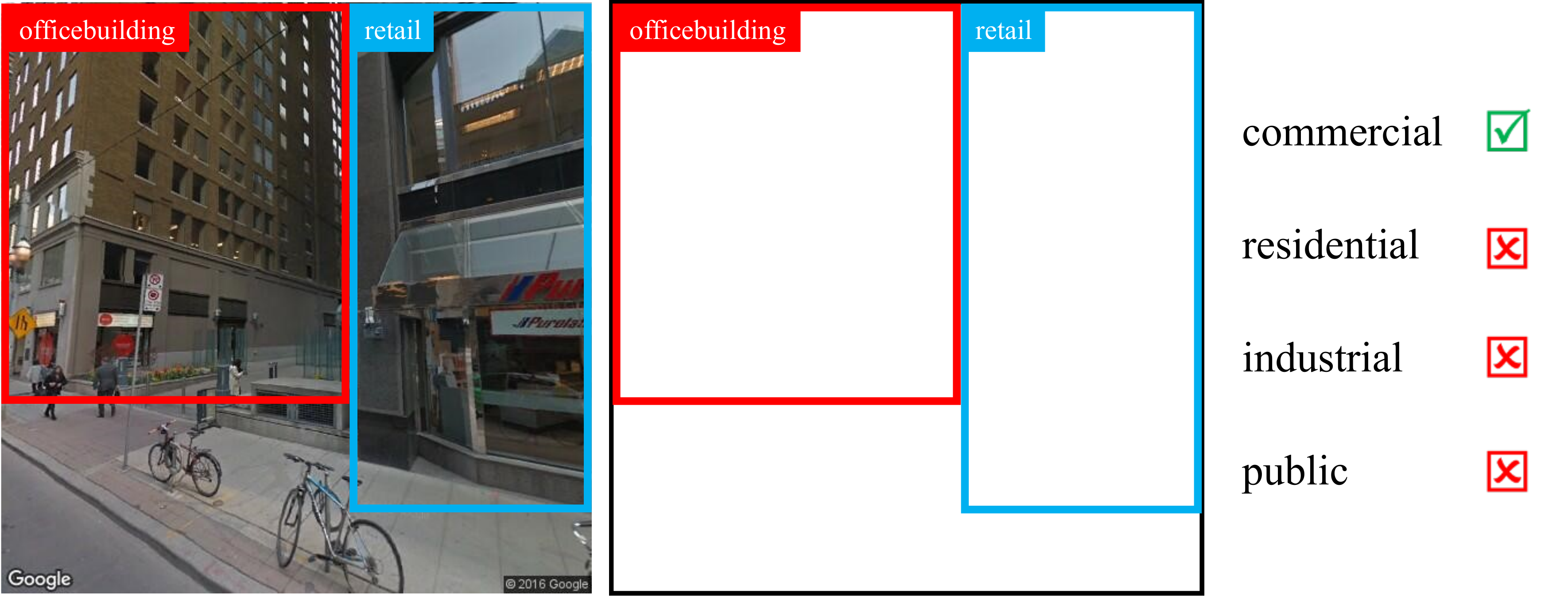} 
			\label{fig:BuildingLabel}      
		\end{minipage}
	}
	\caption{Comparison between common label system and land-use oriented label system in object-level. The subclasses of building enable us to easily distinguish urban functional zones.}
	\label{fig:BoxToScene}
\end{figure}

On the first point, significant objects change with specific tasks. Buildings are the main places where people engage in social and economic activities. Urban functional zones also consist mostly of buildings of different categories. Therefore, individual buildings with fine-grained labels should be considered as “significant objects” in street view images for task of urban land use and functional zone analysis. Fig.~\ref{fig:BoxToScene} demonstrate the importance of “significant objects”. Unfortunately, existing open datasets with outdoor scene for common visual tasks~\cite{russakovsky2015imagenet,lin2014microsoft,zhou2014learning,zhou2017scene} and specific visual tasks (e.g., autonomous driving~\cite{cordts2016cityscapes,neuhold2017mapillary}) are all lack of systematic, fine-grained class definition for buildings. As a milestone work for street view image classification, BIC\_GSV~\cite{kang2018building} classifies individual buildings into 8 categories. However, its image-level annotation may cause ambiguity when a street view image contains multi-class buildings. In fact, currently there is no dataset using object-level annotations of fine-grained multi-class buildings for street view images.

On the second point, the dominant model for street view image classification is CNN. Recent works either use CNNs directly for end-to-end image-level classification from the same source~\cite{srivastava2020fine,kang2018building,ilic2019deep}, or CNNs with a two-stream network structure to fuse visual features of images from different sources\cite{cao2018integrating,zhu2019fine,hoffmann2019model,srivastava2019understanding}. As we mentioned, classifying high-level abstract labels directly using visual features may lead to performance bottlenecks. To break the bottlenecks, not only “visual semantics” is needed, but also “visual syntax”. The former can be obtained by encoding the visual features. The later should be learned from the context releations of formers.

\subsection{Contributions}

The contributions of the paper lie in three aspects as follows.

\begin{itemize}
\item Based on BIC\_GSV, a dual-labeled dataset named \textbf{“BEAUTY”} (\textbf{B}uilding d\textbf{E}tection \textbf{A}nd \textbf{U}rban func\textbf{T}ional-zone portra\textbf{Y}ing) with a 19,070 street view images and 38,857 individual buildings by combining automatic labels acquisition from OpenStreetMap (OSM) and expert annotation. It can be used not only for street view image classification aiming at urban land use analysis, but also for multi-class building detection. We also provide baselines for image classification and object detection running on this dataset.

\item Based on BEAUTY, a \textbf{“Detector-Encoder-Classifier”} framework is proposed to replace the common CNN based architecture. As shown in the right column of Fig.~\ref{fig:BuildingLabel}, without “looking” at the whole image, our approach can infer the land use by only using the bounding boxes of detected buildings. In our approach, object detector is regard as a plug-and-play module that can be arbitrarily replaced, which allows the performance of our approach to easily improve synchronously with the improvement of object detection technology.

\item We explored the effect of co-occurrence pattern of multi-class buildings and, further, their spatial layout on urban functional zone analysis. Based on this, we proposed \textbf{“CODING”} (\textbf{C}ontext enc\textbf{O}ding of \textbf{D}etected build\textbf{ING}s) algorithm to encode the contextual information of bounding boxes into metadata which make it easier to further encoding and classifying using RNN or other models.
\end{itemize}

\subsection{Section Arrangement}

The rest of the paper is organized as follows. In Section~\ref{sec:Related Work}, we review related work on land use classification using street-level images and current research progress on scene context modeling. Section~\ref{sec:Dataset} introduces our dataset “BEAUTY”. The proposed approach is expatiated in Section~\ref{sec:Proposed Approach}. Section~\ref{sec:Experiments and Analysis} shows the experimental setup, results and discussions. Section~\ref{sec:Conclusion and Future Work} concludes the paper.

\section{Related Work}
\label{sec:Related Work}

Urban land use classification has been a growing research field as more data from different sources are available. For example, satellite and aerial images data have been mostly used by the remote sensing community~\cite{pandey2019land}, while street-level images were mainly studied by the computer vision community~\cite{lefevre2017toward}. In the latter, social media images and street view images are the two main sources. Both of them are often referred to a scene analysis problem. In this section, we briefly review the research progress on land use classification using street-level images and context modeling for scene analysis.

\subsection{Land Use Classification Using Social Media Images}
\label{subsec:Land Use Classification Using Social Media Images}

Leung and Newsam~\cite{leung2012exploring} first used social media images from Flickr for land use classification. They used the bag of visual words (BOVW) with a soft-weighing scheme to represent image features and then classified them into 3 categories with support vector machine (SVM). Zhu and Newsam improved Leung’s work by using two groups of Flickr images: indoor and outdoor~\cite{zhu2015land}, and replacing BOVW features with pre-trained CNN features~\cite{zhou2014learning}. Antoniou et al.~\cite{antoniou2016investigating} extracted geo-tagged images from Flickr, Panoramio and Geograph for an area of London, and discussed their usefulness for land use classification. Based on Antoniou’s work, Tracewski et al.~\cite{tracewski2017repurposing} used Places205-AlexNet~\cite{zhou2014learning} to classify social media images with volunteered geographic information (VGI) for land use classification. Zhu et al. coupled images from Google Places and Flickr with a two-stream CNN to predict the land use~\cite{zhu2019fine}. By using ResNet101 as backbone of each branch, they reported 49.54\% classification accuracy on 45 categories. Hoffmann et al.~\cite{hoffmann2019building} classified building instance into 5 land use categories by training a VGG16 using Flickr images.

Social media images provide more street-level details for land use classification. However, they also have shortcomings. First of all, they are often not accurately georeferenced. What’s more, they usually portray highly personalized content (e.g., touristic viewpoints, selfies or zoomed objects) from a subjective, fickle perspective, rather than urban objects from a relatively objective, fixed perspective. Last but not least, they tend to cover the city unevenly (e.g., most images are taken in touristic areas). These problems make such street-level images less suitable for reliable urban land use mapping.

\begin{figure*}[htbp]
	\centering
	\subfigure[Four types of removed samples: indoor, severe occlusion, too large and too small on scale from left to right respectively.]
	{\begin{minipage}[t]{0.8\textwidth}
			\includegraphics[width=\linewidth]{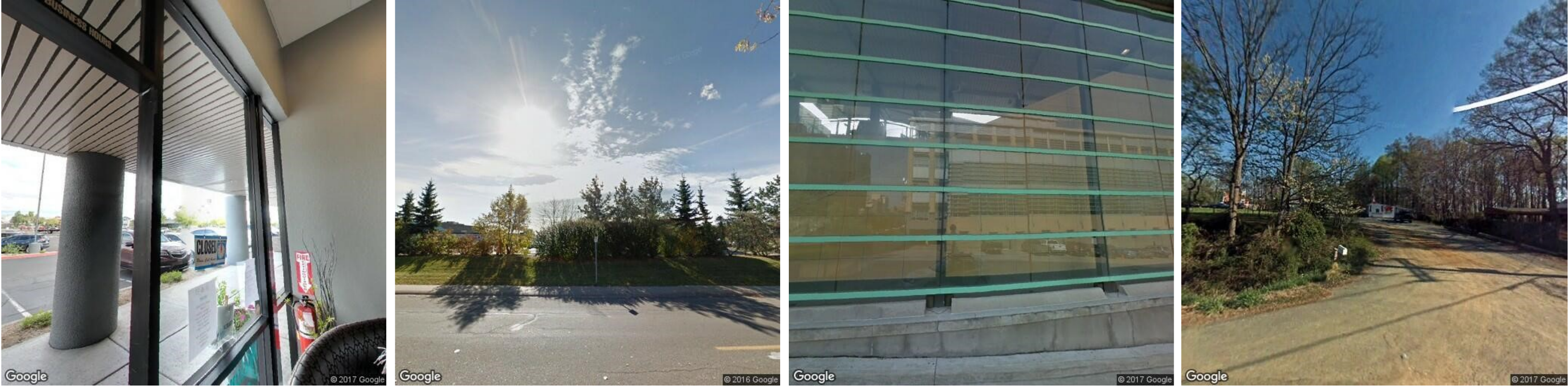} 
			\label{fig:Removed}       
		\end{minipage}
	}
	\subfigure[The comparison of the annotation used in BIC\_GSV~\cite{kang2018building} (left 1, left 2), BIG~\cite{srivastava2018multilabel} (left 3) and BEAUTY (left 4).]
	{\begin{minipage}[t]{0.8\textwidth}
			\includegraphics[width=\linewidth]{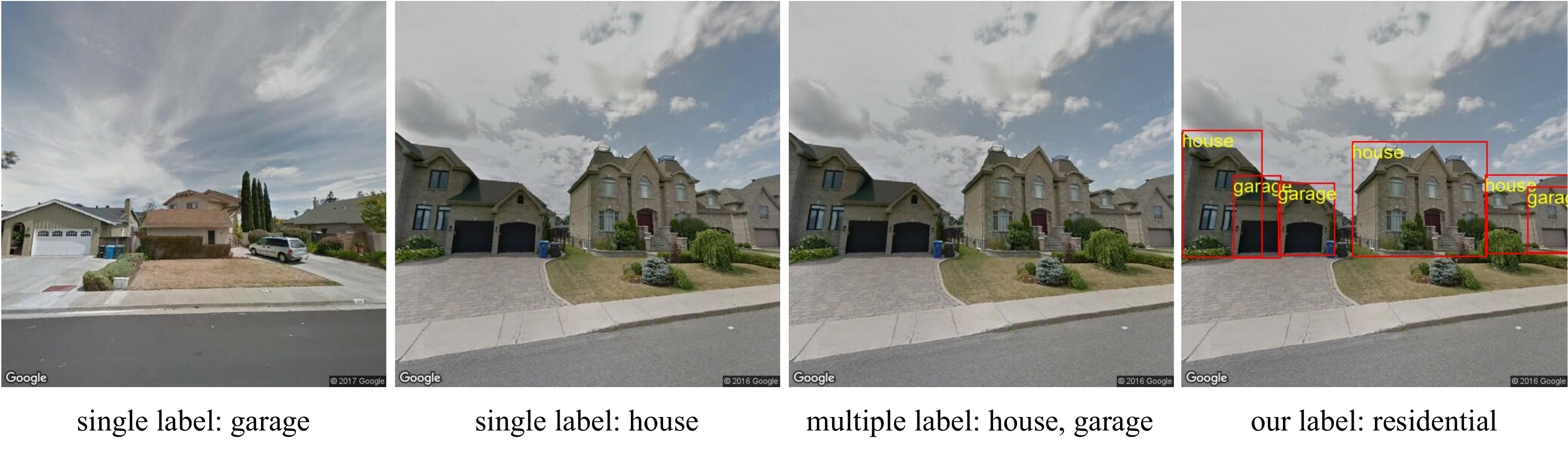} 
			\label{fig:OurLabel}      
		\end{minipage}
	}
	\caption{The improvements made by BEAUTY over BIC\_GSV.}
	\label{fig:BEAUTYvsBIC}
\end{figure*}

\subsection{Land Use Classification Using Street View Images}
\label{subsec:Land Use Classification Using Street View Images}

Services like Google Street View (GSV) make it is possible to acquire street-level images with urban objects shot from a relatively objective perspective, which are accurately geo-located, updated regularly and densely available in many cities all over the world. Recently GSV is being increasingly used in land use classification. Movshovitz et al.~\cite{movshovitz2015ontological} used CNN to classify store fronts into 13 business categories from single GSV images. Kang et al.~\cite{kang2018building} classified urban buildings into 8 categories using GSV images with labels from OSM. Their model predicts one label for each image corresponding to one urban building. Srivastava et al. fused multiple GSV images of a building using a Siamese-like CNN~\cite{srivastava2018multilabel} and showed an overall accuracy of 62.52\% on 16 OSM label prediction. Noticing that each land use category is made of different objects present in a set of images, they then extended their approach to multi-label prediction~\cite{srivastava2020fine} to avoid the ambiguity caused by single-label image classification.

Researchers also try to fuse street view images with overhead images by multi-modal strategies. Combination of both modalities was initially used in image geo-localization. Lin et al.~\cite{lin2013cross} matched HRO from Bing Maps with street view images from Panoramio by using four handcrafted features and adding land cover features as the third modality. To extend their approach, they used a Siamese-like CNN to learn deep features between GSV images and 45-degree oblique aerial images~\cite{lin2015learning}. Workman et al.~\cite{workman2017unified} fused overhead images and GSV images by an end-to-end deep network which outputs a pixel-level labeling of overhead images for three different classification problems: land use, building function and building age. They reported a top-1 accuracy of 77.40\% and 70.55\% for land use classification task on Brooklyn and Queens datasets respectively. Zhang et al.~\cite{zhang2017parcel} combined airborne light detection and ranging (LiDAR) data, HRO and GSV images for land use classification. In their study, thirteen parcel features were chosen as input variables in a Random Forest classifier which achieves an average accuracy of 77.50\%. Cao et al.~\cite{cao2018integrating} used images from Bing Maps and GSV for land use segmentation with a two-stream encoder and one decoder architecture which evolved from SegNet~\cite{badrinarayanan2017segnet}. Hoffmann et al.~\cite{hoffmann2019model} used a two-stream CNN model for building functions classification. They predicted four class labels namely commercial, residential, public and industrial for overhead images by fusing deep features of overhead images and street view images. Their model increases the precision scores from 68\% to 76\% with a decision-level fusion strategy. Srivastava et al.~\cite{srivastava2019understanding} extend their early work~\cite{srivastava2018multilabel} to a multi-modal strategy by leveraging the complementarity of overhead and street-level views. They deal with the situation of missing overhead imagery by using canonical correlation analysis (CCA) based on their two-stream CNN model. By using VGG16 as the backbone, their model achieves an overall accuracy of 73.44\% and an average accuracy of 70.30\%.

Although the usage of multi-modal strategies gets better results to some extent, the performance so far has been less than satisfactory partly due to the high-level abstractness of urban land use labels which were hard to be abstracted directly using visual features. To break the bottlenecks, some new point of view is needed.

\subsection{Context Modeling for Scene Analysis}
\label{subsec:Context Modeling for Scene Analysis}

Image context contains a wealth of information about how objects and scenes are related. Cognitive science studies~\cite{bar2004visual,chun1998contextual} have shown the importance of contextual information in human visual recognition. Typical contextual information including global context~\cite{torralba2010using}, visual context~\cite{dvornik2018modeling}, object co-occurrence~\cite{shih2017deep} and layout~\cite{desai2011discriminative}, are now exploited to improve the performance of various visual tasks. Pathak et al.~\cite{pathak2016context} proposed a context encoder to generate the contents of an arbitrary image region conditioned on its surroundings. Choi et al.~\cite{choi2012context} present a graphical model that combines different sources of context information to detect out-of-context objects and scenes. Izadinia et al.~\cite{izadinia2014incorporating} encoded the scene category, the context-specific appearances of objects and their layouts to learn scene structures. Chien et al.~\cite{chien2017detecting} built a CNN to predict the probability of observing a pedestrian at some location in image. Wang et al.~\cite{wang2017binge} used a variational auto-encoder to extract the scale and deformation of the human pose and thus predict opportunities of interaction in a scene. Qiao et al.~\cite{qiao2019tell} proposed an encoder-generator model that encodes the properties of input objects and generates a scene layout representing the scene context. We consider street view image classification as a fine-grained outdoor scene analysis problem. The proposed context encoder will be detailed in Section~\ref{sec:Proposed Approach}.

\section{Dataset}
\label{sec:Dataset}

As we mentioned in Section~\ref{subsec:Motivation}, currently there is no dataset using object-level annotations of fine-grained multi-class buildings for street view images. Most existing street view datasets use the single-label image-level annotation which contains only global semantics but no descriptions of content or context. Srivastava et al.~\cite{srivastava2018multilabel} used a multi-label image-level annotation dataset namely “BAG”\footnote{\url{https://business.gov.nl/regulation/addresses-and-buildings-databases/}}  which contains object co-occurrence information that could be used to describe contextual relations of the image scene to some extent. However, the labels are for individual buildings such as “office” and “shop”, which lack global semantics of the land use. Furthermore, image-level annotation cannot provide spatial information of objects (e.g., size and position). Thus it contains no richer context information such as layout.

\begin{table*}[htbp]
	\centering
	\caption{The correspondence between the proposed four land use categories and OSM, LBCS labels.}
	\label{tab:Labels}
	\begin{tabular}{lll}
		\Xhline{1.5pt}
		\noalign{\smallskip}
		\textbf{OpenStreetMap Land Use Tag} & \textbf{Proposed Urban Land Use Classes} & \textbf{LBCS Function Dimension} \\
		\Xhline{1pt}
		\noalign{\smallskip}
		residential & residential & 1000: residence or accommodation functions \\
		\noalign{\smallskip}
		garages & residential & - \\
		\noalign{\smallskip}
		commercial & commercial & 2000: general sales or services \\
		\noalign{\smallskip}
		retail & commercial & 2000: general sales or services \\
		\noalign{\smallskip}
		cemetery & public & - \\
		\noalign{\smallskip}
		recreation ground & public & 6000: education, public admin., health care, religious and other institutions \\
		\noalign{\smallskip}
		religious & public & 6000: education, public admin., health care, religious and other institutions \\
		\noalign{\smallskip}
		village green & public & 6000: education, public admin., health care, religious and other institutions \\
		\noalign{\smallskip}
		- & public & 4000: transportation, communication, information, and utilities \\
		\noalign{\smallskip}
		industrial & industrial & 3000: manufacturing and wholesale trade \\
		\noalign{\smallskip}
		\Xhline{1.5pt}
	\end{tabular}
\end{table*}

To explore the context relations between street view scene and urban objects in it, a street view image dataset with a dual-label system is made based on the existing BIC\_GSV dataset~\cite{kang2018building}. On one hand, each image has a land use label to describe the functional zone it portrays, such as “commercial”. On the other hand, each urban object (mostly individual building) in the image is annotated by a bounding box with an object-level label such as “retail”. Thus, the proposed dataset named \textbf{“BEAUTY”}  can be use both in land use classification task and in individual building detection task.

BIC\_GSV obtained geo-tagged GSV images located over several cities of the US and Canada (e.g., Montreal, New York City and Denver) and their associated ground truth building labels extracted from OSM. BEAUTY makes the following improvements over BIC\_GSV.

\begin{itemize}
	\item The remaining invalid samples are further removed. Although BIC\_GSV has removed some outliers with VGG16 trained on Places2~\cite{zhou2014learning}, some invalid samples were still found during the manual inspection. As shown in Fig.~\ref{fig:Removed}, we further remove four types of remaining invalid samples: indoor, severe occlusion, too large and too small on scale.
\end{itemize}

\begin{itemize}	
	\item Object-level annotations are given for each building in an image. In combination with the building labels\footnote{\url{https://wiki.openstreetmap.org/wiki/Map_Features\#Building}} automatically obtained from OSM, we manually annotate each individual building in each image under the guidance of architecture experts. In object-level annotations, we use the 8 class labels used in BIC\_GSV, namely \textit{apartment}, \textit{church}, \textit{garage}, \textit{house}, \textit{industrial}, \textit{office building}, \textit{retail} and \textit{roof}. Object-level annotations avoid the ambiguity when buildings in different classes are in the same image and also afford the layout information of buildings in the same scene of land use. An example is shown in Fig.~\ref{fig:OurLabel}.
\end{itemize}

\begin{itemize}		
	\item Image-level labels are further abstracted into land use categories. In combination with the land use labels\footnote{\url{https://wiki.openstreetmap.org/wiki/Map_Features\#Landuse}} automatically obtained from OSM and the Land Based Classification Standards (LBCS) Function Dimension with Descriptions\footnote{\url{https://www.planning.org/lbcs/standards/function/}}, we manually annotate each image under the guidance of urbanist. We fuse the OSM land use labels and LBCS urban function descriptions into 4 highly compact classes namely \textit{commercial}, \textit{residential}, \textit{public} and \textit{industrial}, which have been used in~\cite{hoffmann2019model}, because such a classification has a very high value to urban geography being correlated with socio-demographic parameters such as population density and income. The correspondence between the proposed four land use categories and OSM, LBCS labels is shown in TABLE~\ref{tab:Labels}.
\end{itemize}

\begin{figure}[h]
	\centering
	\subfigure[Sample numbers of each land use class.]
	{\begin{minipage}[t]{0.45\textwidth}
			\includegraphics[width=\linewidth]{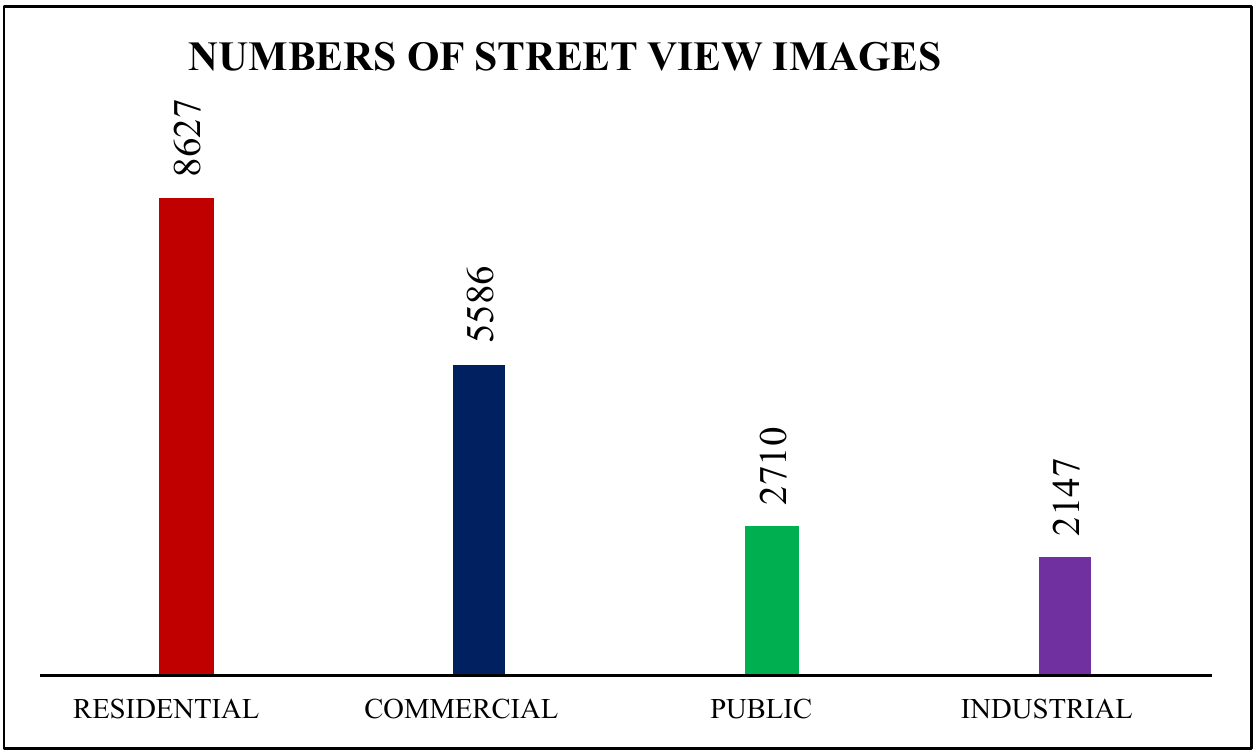}     \label{fig:BeforeCl}   
		\end{minipage}
	}
	\subfigure[Sample numbers of each building class.]
	{\begin{minipage}[t]{0.45\textwidth}
			\includegraphics[width=\linewidth]{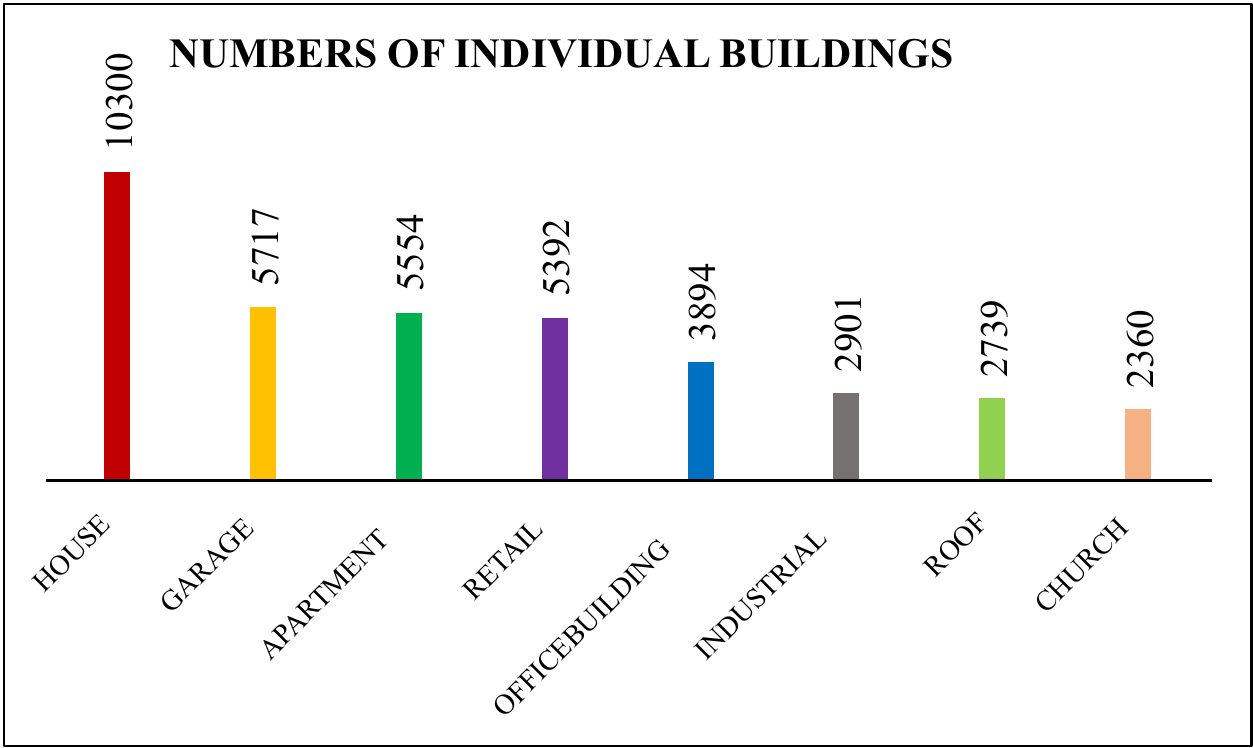}     \label{fig:BeforeDe}
		\end{minipage}
	}
	\caption{Sample distributions of BEAUTY.}
	\label{fig:BeforeReb}
\end{figure}

\begin{figure*}[!htb]
	\centering
	\includegraphics[width=0.7\linewidth]{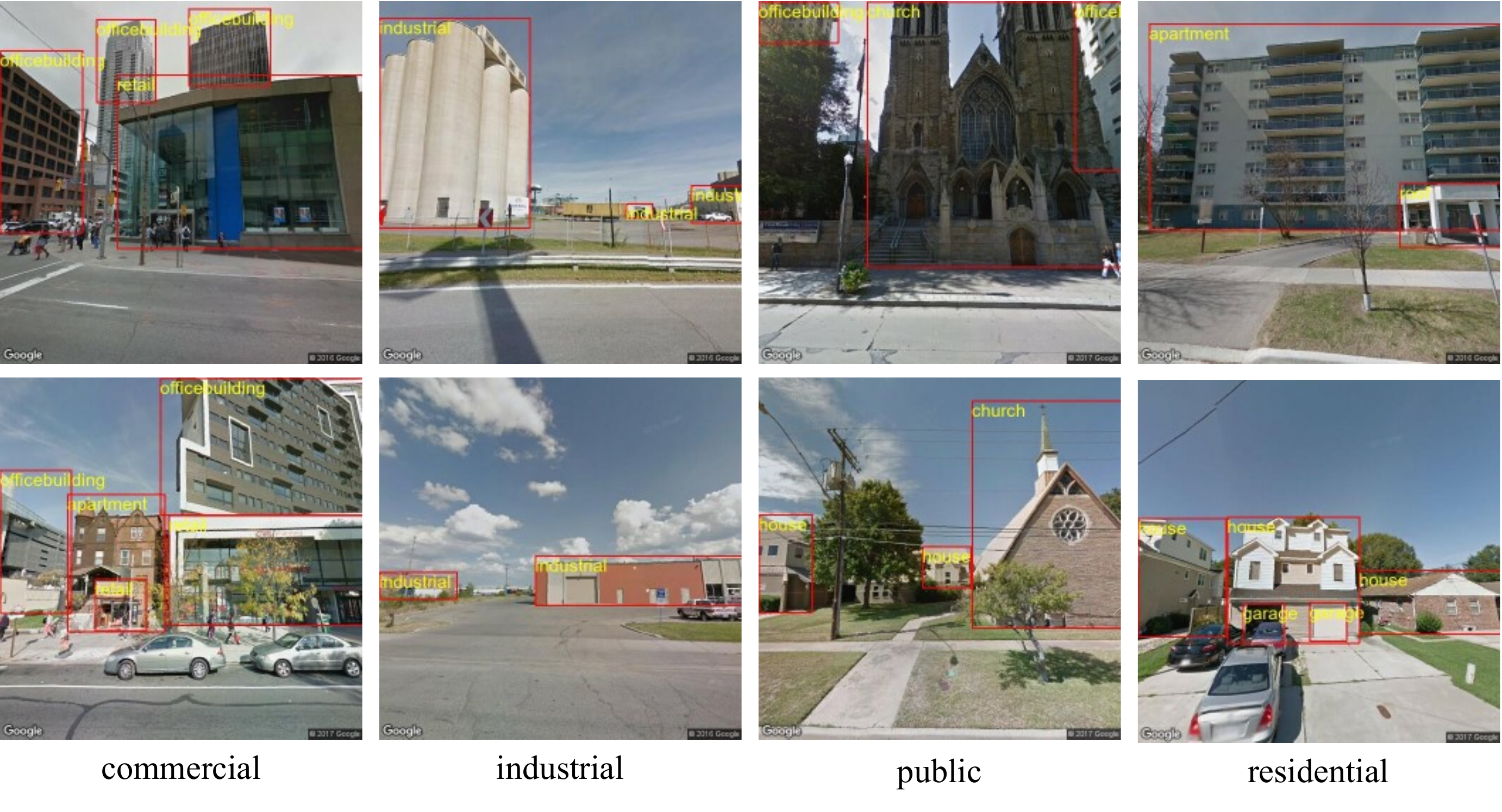}
	\caption{Samples of BEAUTY: street view images in four land use scenes with different type of buildings.}
	\label{fig:Samples}
\end{figure*}

The BEAUTY dataset consists of 19,070 street view images with 38,857 individual buildings. As can be seen from Fig.~\ref{fig:BeforeReb}, both the sample distributions of land use classes and building classes are long-tailed, which are in line with the situation in the real world. Fig.~\ref{fig:Samples} shows samples of proposed dataset.

\section{Proposed Approach}
\label{sec:Proposed Approach}

\begin{figure*}[!htb]
	\centering
	\includegraphics[width=0.8\linewidth]{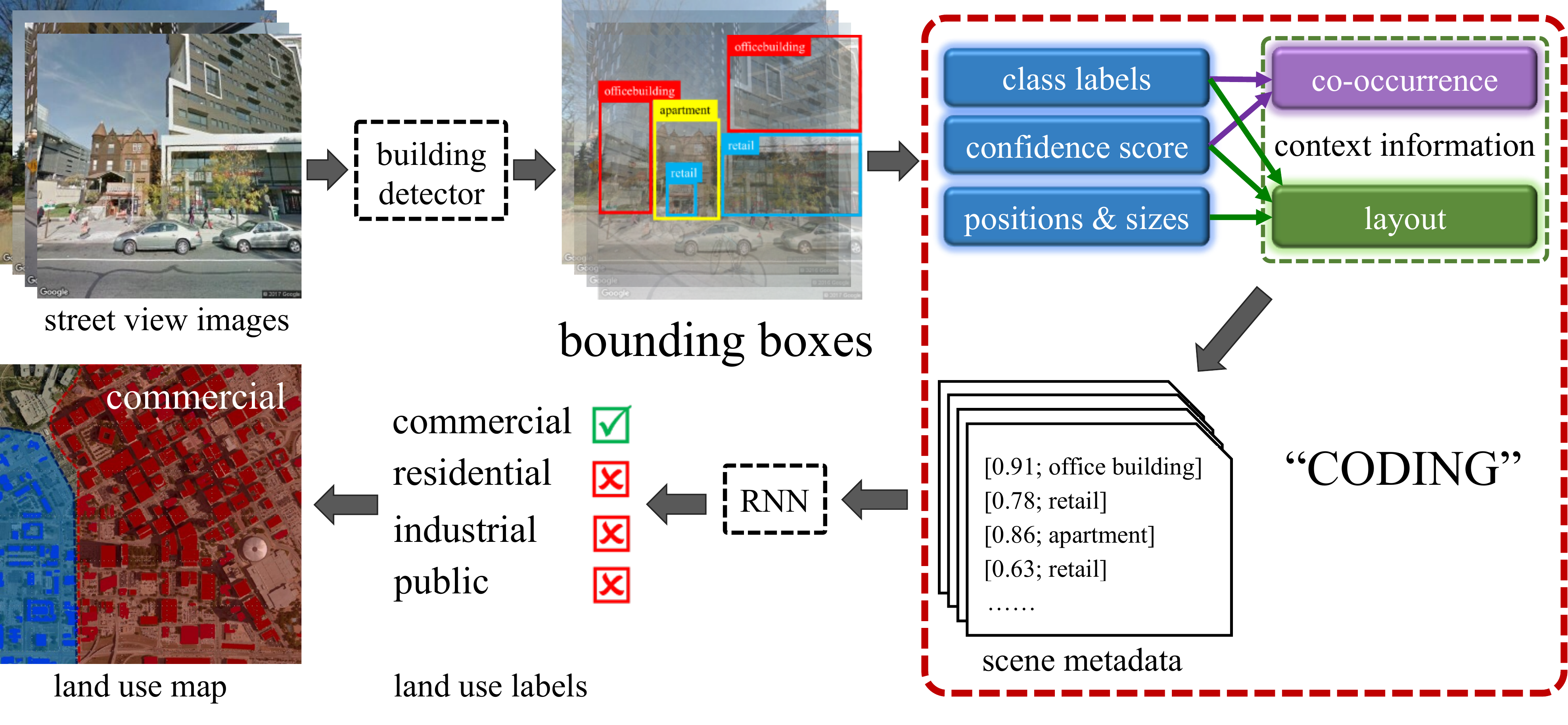}
	\caption{Pipeline of proposed approach using a “Detector-Encoder-Classifier” framework. The core algorithm “CODING” encodes the input bounding boxes into scene metadata containing contextual information.}
	\label{fig:PipeLine}
\end{figure*}

Fig.~\ref{fig:PipeLine} shows the pipeline of proposed approach. The inputs are street view images and the outputs are their predicted land use categories which would be mapped to the geographic information systems according to their geo-location.

As we mentioned in Section~\ref{subsec:Motivation}, the first key of our task is acquiring the most significant urban objects in street view images. The building detector plays the key role to do it. Two off-the-shelf detectors were used in this paper namely Faster R-CNN~\cite{ren2017faster} and Cascaded R-CNN~\cite{cai2019cascade}. The detectors were trained using the object-level ground truth of training samples in BEAUTY. The outputs of the detector are bounding boxes of each building with their classes and confidence scores, which would be transferred into metadata through the context encoder. In addition to being an intermediate module of the street view classification task, we also consider building detection as a separate task and conduct the corresponding baseline tests. This part will be detailed in Section~\ref{subsec:Baselines}.

As shown in Fig.~\ref{fig:PipeLine}, two different kinds of contextual information are optional in the proposed “CODING” module, which are co-occurrence patterns and layout.

\subsection{Context Encoding Using Object Co-occurrence Only}
\label{subsub:Encoding Using Object Co-occurrence Only}

Outputs of a detector are bounding boxes of detected object regions, each of which consists of the following data: predicted confidence score, class label and position in the image. The class and confidence score are concrete representations of semantic information. We integrate them together into a feature vector with the form of the hot-one vector of predicted class whose none-zero value was replaced by the confidence score. These “semantic vectors” are used directly to the classification task. The position data are often with the form of $[x_i/W, y_i/H, w_i/W, h_i/H]$ where $x_i$, $y_i$ are the coordinates of the top left corner of the detection bounding box, $w_i$, $h_i$ are its width and height, and $W$, $H$ are the width and height of the image. The use of position information will be explained in detail in the next part.

The semantic vectors of detections are grouped by image and mapped to a set which is padded to length $l$ by all-zero vectors. We set $l = M + m$, where $M$ is the max detections of one image in training set and $m$ is a slack. A set of vectors is obtained without using position data. The set contains only co-occurrence information of different buildings in a scene and could be further encoded and classified.

\subsection{Context Encoding Using Building Layout}
\label{subsub:Encoding Using Building Layout}

The absolute position of bounding boxes were not encoded directly because the angle and scale of the building shot in the street view image change dramatically, which makes the features lack of angle and scale invariance. The position vectors are used to compute intermediate variables such as the relative size of bounding boxes and the distance between them. These intermediate variables help to decide the order of semantic vectors. Thus, the sequence implies the relative layout of buildings and preserves the invariance of angle and scale of features simultaneously. The specific steps of sequence generation are shown in Algorithm~\ref{alg:layout}, where $\mathbb{B}$ denotes the set of detected bounding boxes $B_i$ with its hot-one vector of class $\vec{C}_i$
, predicted confidence score $p_i$ and its position vector $[x_i, y_i, w_i, h_i]$, and $\mathbb{S}$ denotes the generated sequence.

\begin{algorithm}[htb]
	%\setstretch{1.2}
	\caption{Sequence generation for layout encoding}
	\label{alg:layout}
	\begin{algorithmic}[1]
		\STATE{\textbf{Input:} Set of bounding boxes $\mathbb{B}$}
		\STATE{\textbf{Output:} Sequence of semantic vectors $\mathbb{S}$}
		\STATE{$\mathbb{S} \leftarrow \emptyset$} 
		\FOR{$B_i \in \mathbb{B}$}
		\STATE{$a_i \leftarrow w_i \times h_i \times p_i$} \label{alg:times}
		\STATE{$\hat{x}_i \leftarrow x_i + w_i/2$}
		\STATE{$\hat{y}_i \leftarrow y_i + h_i/2$}
		\STATE{$\vec{C}_i^* \leftarrow p_i \times \vec{C}_i$}
		\ENDFOR
		\STATE{PUSH $\vec{C}_0^* : a_0 = \max \{a_i\}$ INTO $\mathbb{S}$}
	    \label{alg:alpha}
		\STATE{DELETE $B_0 : a_0 = \max \{a_i\}$ FROM $\mathbb{B}$}
		\STATE{DELETE $\vec{C}_0^* : a_0 = \max \{a_i\}$ FROM $\{\vec{C}_i^*\}$}
		\FOR{$B_i \in \mathbb{B}$}
		\STATE{$d_i \leftarrow \sqrt{(\hat{x}_i - \hat{x}_0)^2 + (\hat{y}_i - \hat{y}_0)^2}$}
		\label{alg:distance}
		\ENDFOR
		\STATE{ASCENDING\_SORT $\vec{C}_i^*$ BY $d_i$}
		\label{alg:sort}
		\FOR{$\vec{C}_i^* \in \{\vec{C}_i^*\}$}
		\STATE{PUSH $\vec{C}_i^*$ INTO $\mathbb{S}$}
		\ENDFOR
		\STATE{RETURN $\mathbb{S}$}
	\end{algorithmic}
\end{algorithm}

To put it simply, we first select the bounding box with highest confidence score and largest size (Line~\ref{alg:times}) to be the leading box (Line~\ref{alg:alpha}), and then ascending sort the rest ones (Line~\ref{alg:sort}) by the centroids distance (Line~\ref{alg:distance}) between them and the leading box. Finally, sequences of vectors with hot-one like form (semantic structure), and their order (syntax structure) constitute the scene metadata.

To further encoding and classifying the metadata obtained from “CODING”, two RNN architectures are used, namely last-layer-concatenated single-directional RNN and all-concatenated bidirectional RNN (BRNN)~\cite{schuster1997bidirectional}, both with two hidden layers. The inputs of RNNs are the semantic vectors with the size of 8, representing the detected bounding boxes of 8 class buildings in an image. The size of a hidden layer neuron $h_i$ is 16 and the dimension of the parameter matrix $\mathcal{A}$ in the first hidden layer is $16 \times (16+8)$. The basic units follow a simple RNN structure. It can also be replaced by a gated recurrent unit (GRU)~\cite{cho2014properties} or a long short-term memory (LSTM)~\cite{hochreiter1997long} unit, which will be compared in Section~\ref{subsec:Comparison of Different Settings}.

In the first architecture, all hidden neurons in the last layer are concatenated. In order to reduce the weights of zero vectors generated by padding, vectors of the input metadata should be arranged in reverse order. The concatenation layer ensures that no feature of a single bounding box would be forgotten by the directionality of the RNN. Well in the BRNN architecture, all neurons in both hidden layers are finally concatenated to connect to a full connection (FC) layer and output the predicted probability of four classes of land use scenes after a softmax operation.

\section{Experiments and Analysis}
\label{sec:Experiments and Analysis}

In order to verify the validity of the proposed approach, we ran it on BEAUTY dataset which has been introduced in detail in Section~\ref{sec:Dataset}. In this section, we will first conduct baseline tests for the tasks of street view image classification and building detection on this dataset, then compare the performance of proposed approach with baseline, and draw some useful conclusions through analysis.

\begin{figure}[h]
	\centering
	\subfigure[Sample numbers of each land use class.]
	{\begin{minipage}[t]{0.45\textwidth}
			\includegraphics[width=\linewidth]{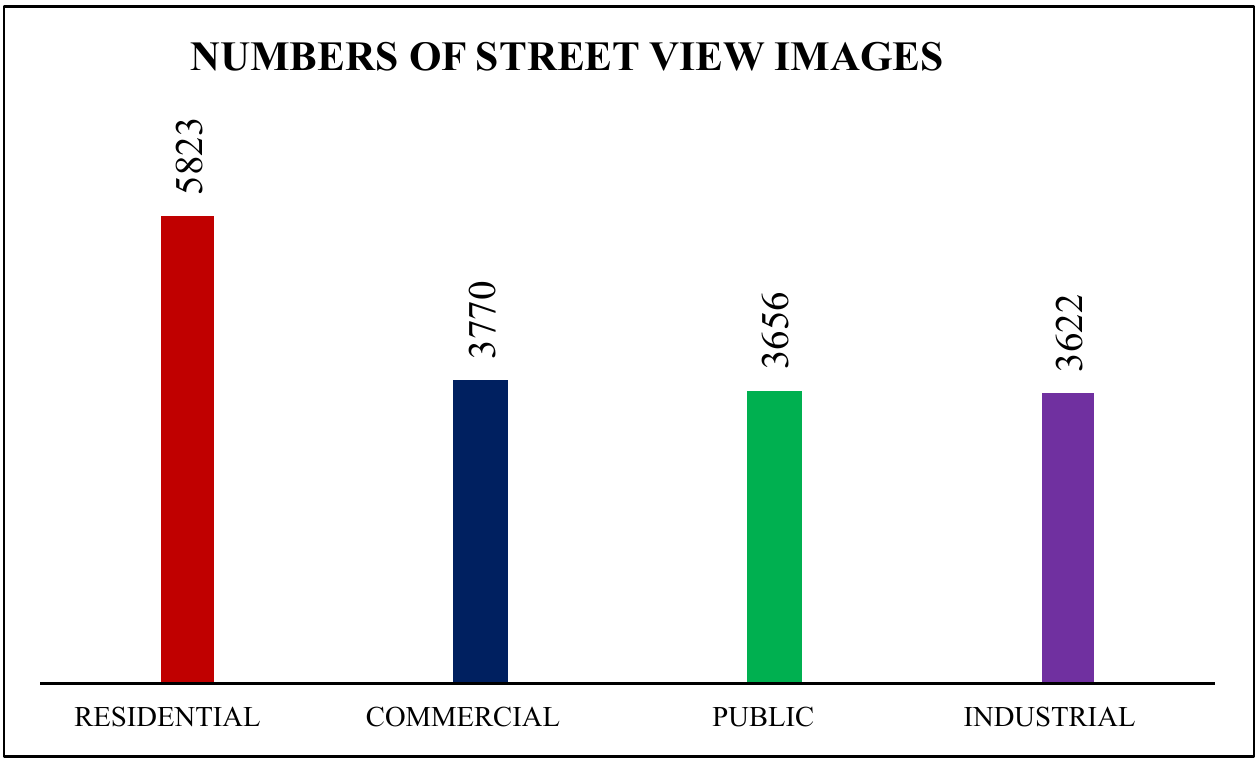}     \label{fig:AfterCl}   
		\end{minipage}
	}
	\subfigure[Sample numbers of each building class.]
	{\begin{minipage}[t]{0.45\textwidth}
			\includegraphics[width=\linewidth]{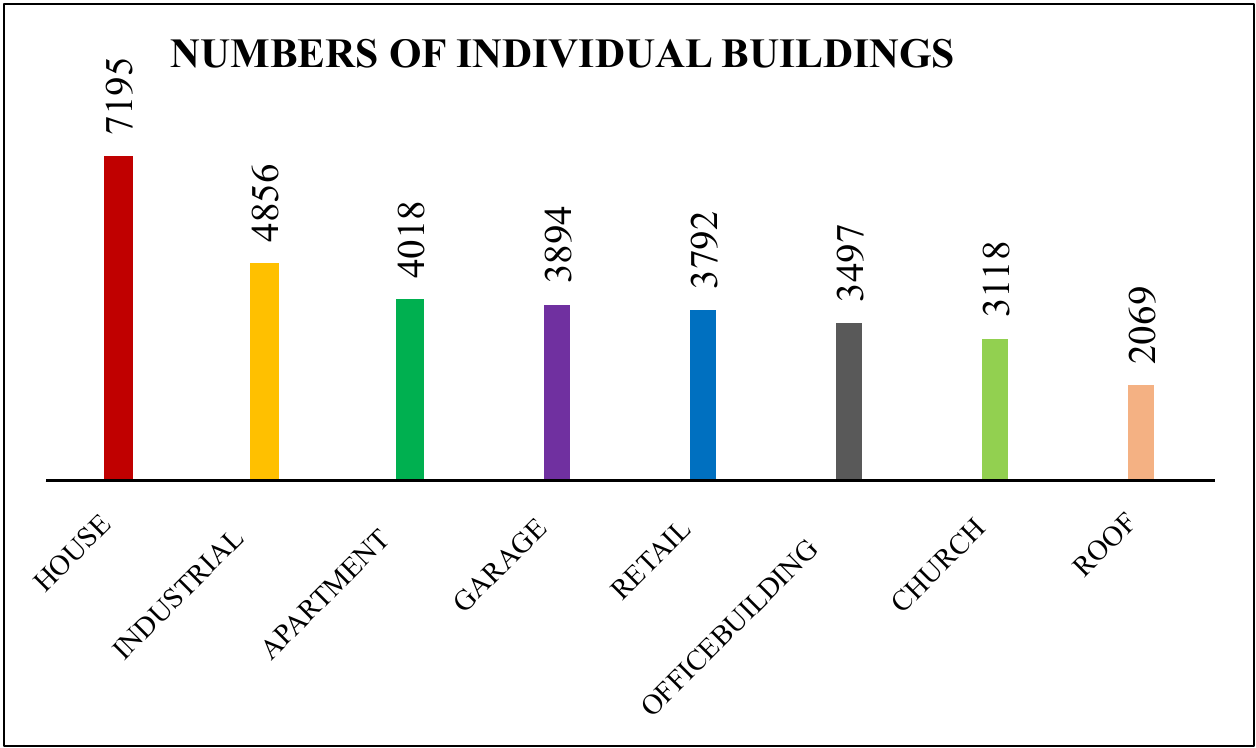}     \label{fig:AfterDe}
		\end{minipage}
	}
	\caption{Training sample distributions of land use classes and building classes after rebalance.}
	\label{fig:AfterReb}
\end{figure}

\subsection{Data Preparation and Experimental Setup}
\label{subsec:Data Preparation and Experimental Setup}

We randomly selected 75\% of the samples from each category as the training/validation set and the remaining 25\% as the test set. The training/verification set was then randomly divided into training set and verification set according to the ratio of 9:1. From Fig.~\ref{fig:BeforeReb} we can learn that there is a class imbalance problem~\cite{buda2018systematic} in BEAUTY. To reduce the impact of class imbalance and achieve better performance, we carried out class rebalance for the training samples using a random minority oversampling strategy. Specifically, the samples of \textit{public} and \textit{industrial} were expanded by 2 times and 2.5 times respectively. Since all training samples will be flipped randomly in the horizontal direction in the data augmentation stage before training, only random copy is needed for minority oversampling, which is a common strategy in the industry. We also tried random geometry transformation and random color jittering~\cite{zhang2019bag} for training data augmentation. Unfortunately, the performance is not as good as random horizontal flip considering both the tasks of street view image classification and building detection. The sample distributions after rebalance are shown in Fig.~\ref{fig:AfterReb}. The total number of training images and individual buildings are 16,871 and 32,439 respectively after class rebalance before data augmentation. Compare Fig.~\ref{fig:AfterReb} with Fig.~\ref{fig:BeforeReb} we can learn that not only the image-level samples of land use classes are rebalanced, but also the object-level samples of building classes.

All experiments are based on the same hardware and software conditions as follows. GPU: GeForce GTX 1080 $\times$ 2; OS: Ubuntu 18.04.3 LTS; CUDA Version: 10.0.130; PyTorch Version: 1.4.0 for cu100; TorchVision Version: 0.5.0 for cu100. We set the number of RNN input $l=25$, which ensures that most bounding boxes are involved and avoids too much zero vectors. The pre-trained models and the training hyperparameters will be presented in detail in later sections. All the results were averaged after 10 runs.

\subsection{Baselines}
\label{subsec:Baselines}

To facilitate the evaluation of model performance on street view image classification and building detection on BEAUTY, we selected the corresponding baseline models for both tasks. Considering the leading role of CNN-based end-to-end model in image classification and object detection tasks in recent years, we chose the most representative and most widely used ResNet~\cite{he2016deep} model and two detectors based on it as the baseline models.

\subsubsection{Baseline Test for Street View Image Classification}
\label{subsub:Baseline Test for Street View Image Classification}

\begin{figure}[!htb]
	\centering
	\includegraphics[width=0.9\linewidth]{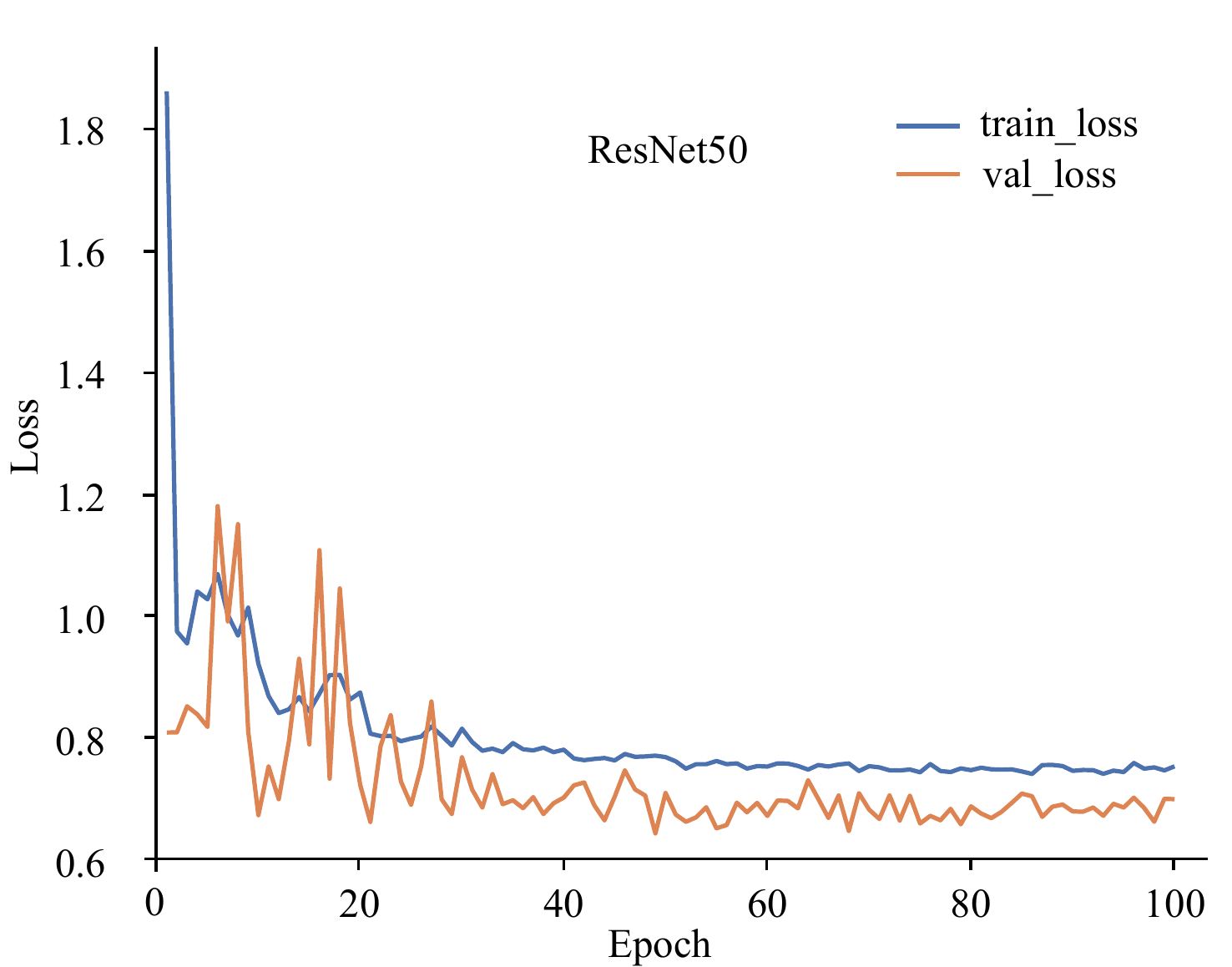}
	\includegraphics[width=0.9\linewidth]{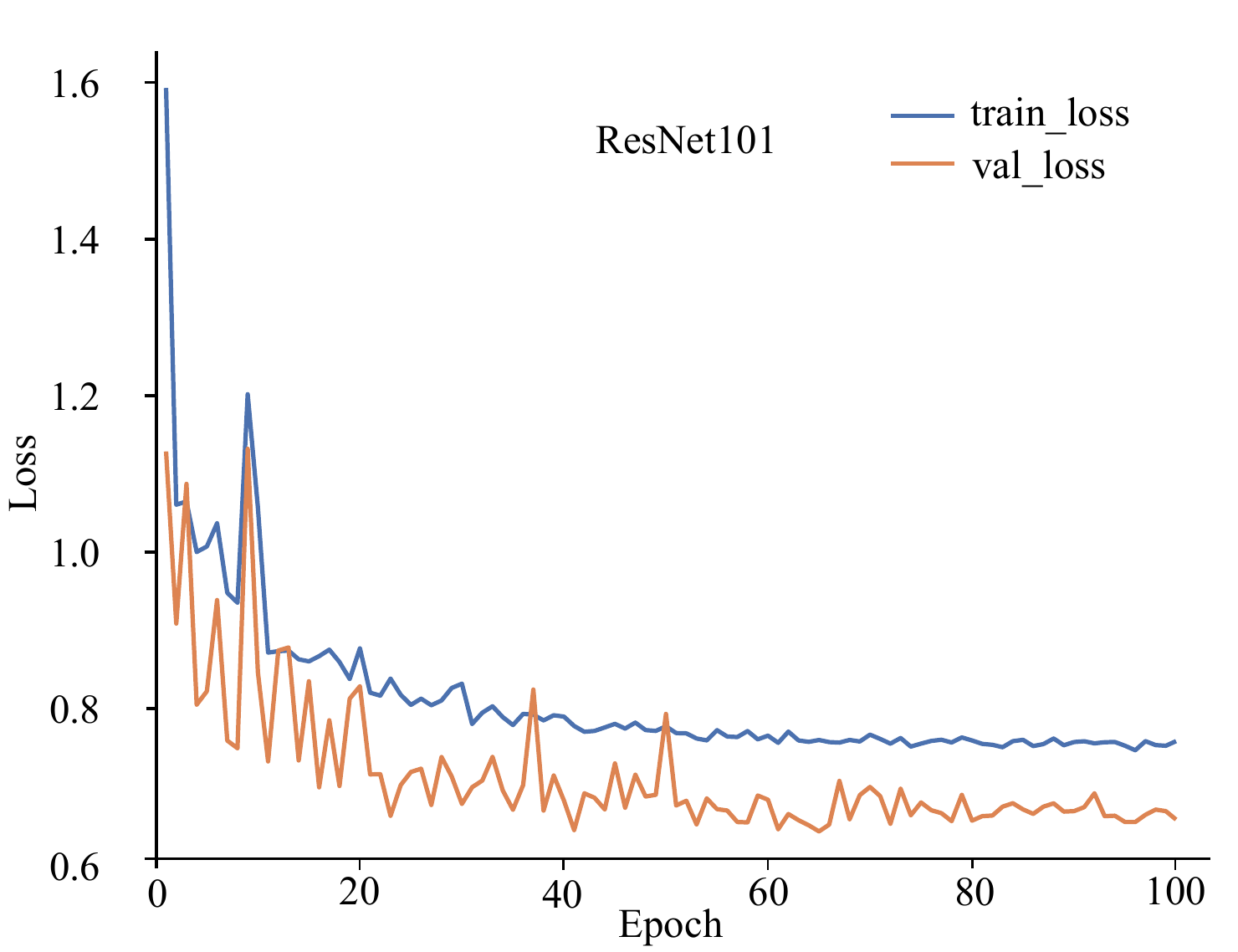}
	\caption{The training and validation losses of ResNet50 and ResNet101.}
	\label{fig:ResNetLoss}
\end{figure}

For the task of street view image classification, ResNet50 and ResNet101 are selected as the candidate baseline models. We finetuned the pre-trained models\footnote{\url{https://pytorch.org/docs/stable/torchvision/models.html}} on BEAUYT with the learning rate of 0.01, which was multiplied by a factor of 0.1 after every 10 epochs. The training was pursued for 100 epochs with Adam~\cite{kingma2014adam} as an optimizer. The training and validation losses are drawn in Fig.~\ref{fig:ResNetLoss}.

\begin{figure*}[!htb]
	\centering
	\includegraphics[width=0.35\linewidth]{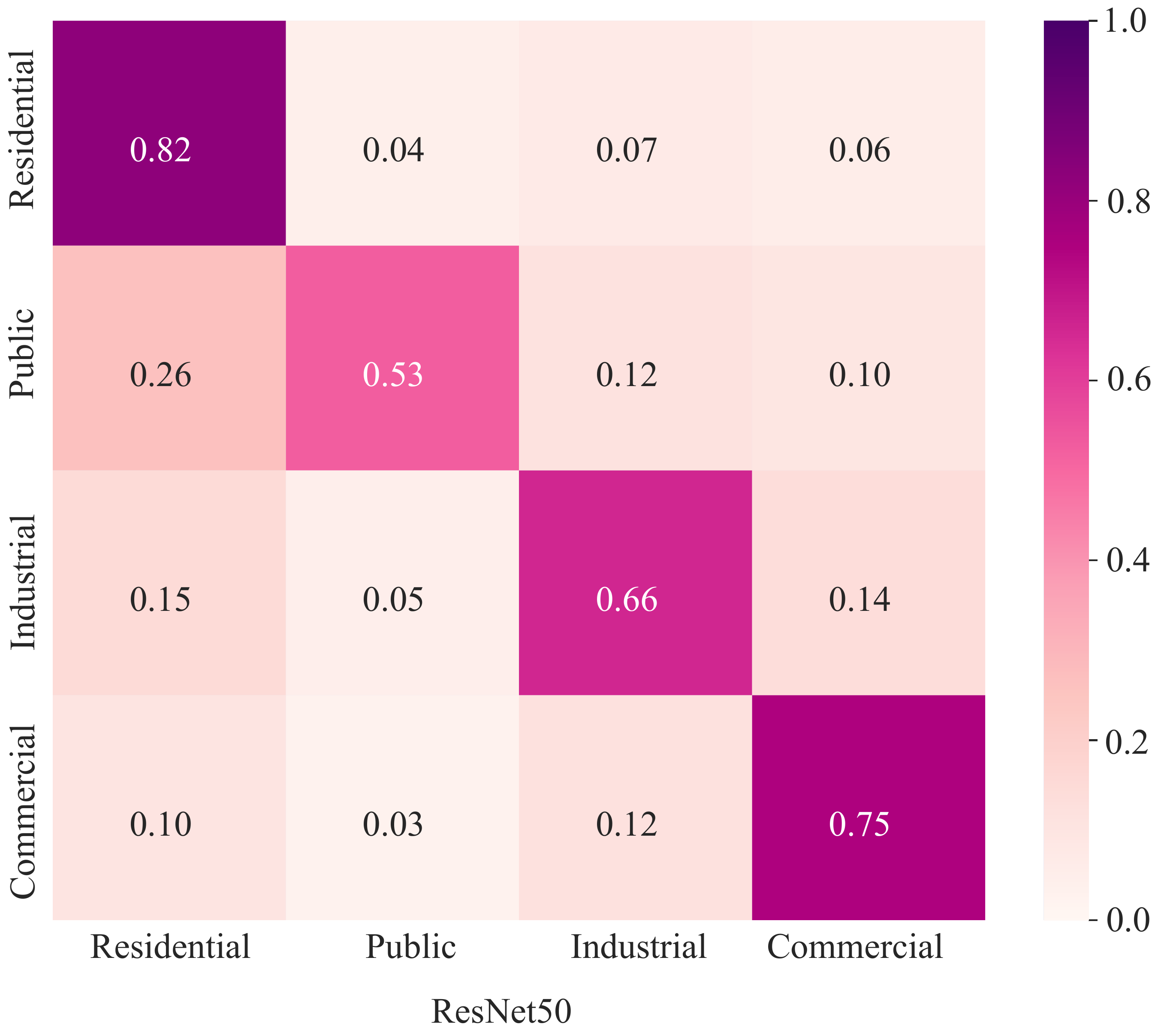}
	~~~~~~~~\includegraphics[width=0.35\linewidth]{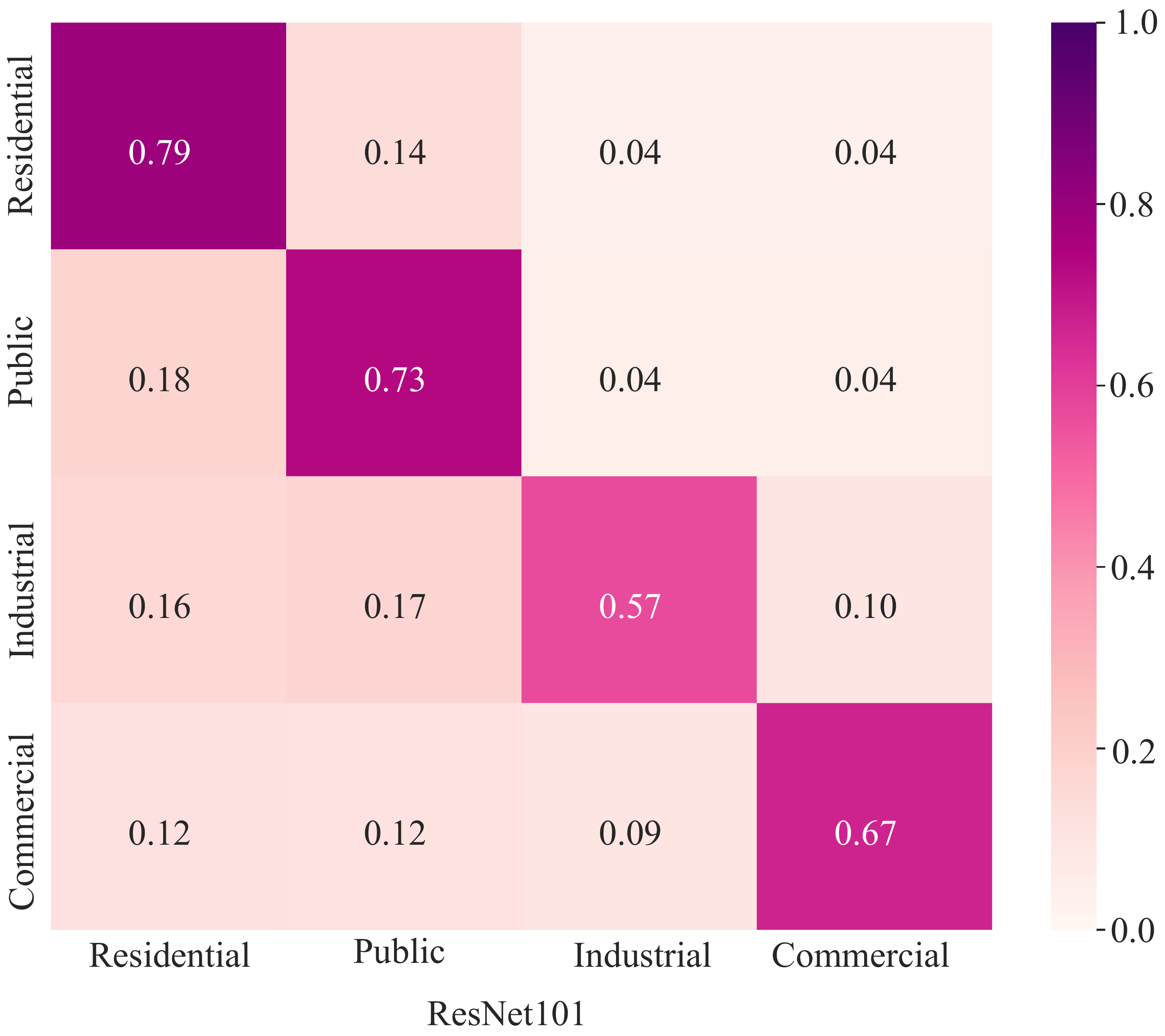}
	\caption{The confusion matrices of ResNet50 and ResNet101.}
	\label{fig:ResNetCon}
\end{figure*}

Although the training set has been rebalanced, serious class imbalance remained in the test set of BEAUTY. Therefore, as a commonly used metric for classification, overall accuracy is not suitable for the evaluation on our dataset. Instead, the macro-average of the per-class metrics are used, namely macro-precision (M-P), macro-recall (M-R) and macro F1-Score (M-F1)~\cite{grandini2020metrics}. As shown in TABLE~\ref{tab:ClassBaseline}, the performance of ResNet50 beats ResNet101 in all macro-average metrics. Thus, ResNet50 is selected as the baseline model for street view image classification on BEAUTY. The confusion matrices in Fig.~\ref{fig:ResNetCon} also show that ResNet50 performs better than ResNet101 in all other categories with the exception of \textit{public}.

\begin{table}[htbp]
	\centering
	\caption{Performances of ResNet50 and ResNet101 in percentage terms.}
	\setlength{\tabcolsep}{20pt}
	\label{tab:ClassBaseline}
	\begin{tabular}{cccc}
		\Xhline{1.5pt}
		\noalign{\smallskip}
		\textbf{Models} & \textbf{M-P} & \textbf{M-R} & \textbf{M-F1} \\
		\Xhline{1pt}
		\noalign{\smallskip}
		ResNet50 & \textbf{69.16} & \textbf{68.94} & \textbf{69.05} \\
		\noalign{\smallskip}
		ResNet101 & 67.48 & 68.87 & 68.17 \\
		\noalign{\smallskip}
		\Xhline{1.5pt}
	\end{tabular}
\end{table}

\subsubsection{Baseline Test for Building Detection}
\label{subsub:Baseline Test for Building Detection}

For the task of building detection, Faster R-CNN~\cite{ren2017faster} and Cascaded R-CNN~\cite{cai2019cascade} with the backbone of ResNet50 and ResNet101 are selected as the candidate baseline models. We finetuned the pre-trained models\footnote{\url{https://mmdetection.readthedocs.io/en/latest/model_zoo.html}} of MMDetection~\cite{chen2019mmdetection} using their default hyperparameters on BEAUTY. Part metrics used in COCO 2020 Object Detection Task\footnote{\url{https://cocodataset.org/\#detection-eval}} are used and extended as our metrics for building detection, which are “average precisions” (AP) at “intersection over union” (IoU) of different values. For example, $\rm AP^{IoU=.50:.05:.95}$ refers to the AP of all classes when the detections and ground truth bounding boxes were matching according to the least IoU value to be 0.5 to 0.95. It is a relatively strict metric, because a high IoU lower limit represents a high requirement for the position accuracy of the prediction box. In contrast, $\rm AP^{IoU=.50}$ is a metric with relatively loose requirement for position accuracy. In order to compare the effectiveness of visual feature extraction between the detector and the end-to-end classifier, we extend this set of metrics to $\rm AP^{IoU=.00}$, which means that the location of an object detection is not considered, but only whether its category is correct for evaluation. In essence, this metric is equivalent to the macro-precision of a multi-label classifier. 

The performances of candidate detectors with candidate backbones on $\rm AP^{IoU=.50:.05:.95}$ to $\rm AP^{IoU=.00}$ are shown in TABLE~\ref{tab:DetectionBaseline}, where Fa-50 refers to Faster R-CNN with the backbone of ResNet50, and Ca-101 refers to Cascaded R-CNN with the backbone of ResNet101. As the metrics became looser, the detectors scored higher. When consider the accuracy of bounding box position, Ca-101 was the best detector. In contrast, Fa-50 becomes optimal when only the accuracy of the class prediction is considered. We select Ca-101 to be the default detector for our system because our approach encodes the layout of buildings by using both the class and position information. If not specified, the detectors used in the subsequent experiments are all Ca-101. TABLE~\ref{tab:DetectionBaseline} is also considered as the baseline for an independent visual task namely multi-class building detection on BEAUTY.

\begin{table}[htbp]
	\centering
	\caption{Performances of candidate detectors in percentage terms.}
	\setlength{\tabcolsep}{4pt}
	\label{tab:DetectionBaseline}
	\begin{tabular}{ccccc}
		\Xhline{1.5pt}
		\noalign{\smallskip}
		\textbf{Detectors} & \textbf{$\rm AP^{IoU=.50:.05:.95}$} & \textbf{$\rm AP^{IoU=.75}$} & \textbf{$\rm AP^{IoU=.50}$} & \textbf{$\rm AP^{IoU=.00}$} \\
		\Xhline{1pt}
		\noalign{\smallskip}
		Fa-50 & 46.09 & 50.71 & 69.70 & \textbf{79.33} \\
		\noalign{\smallskip}
		Fa-101 & 46.11 & 50.73 & 69.42 & 79.01 \\
		\noalign{\smallskip}
		Ca-50 & 48.72 & 53.24 & \textbf{70.21} & 79.11 \\
		\noalign{\smallskip}
		Ca-101 & \textbf{48.92} & \textbf{53.88} & 70.13 & 79.10 \\
		\noalign{\smallskip}
		\Xhline{1.5pt}
	\end{tabular}
\end{table}

Compared with the best M-P (69.16) in TABLE~\ref{tab:ClassBaseline}, the best AP (79.33) in TABLE~\ref{tab:DetectionBaseline} is significantly improved with the same backbone CNN architecture (ResNet50). Although this comparison is not rigorous, we can still roughly observe that the individual buildings in images are easier to be visually characterized and abstracted more effectively than the whole street view image by the same visual feature extractor. This conclusion is the cornerstone of the superiority of our approach over image-level end-to-end CNN models. More details about the effectiveness of visual extraction are presented in Section~\ref{subsub:Typical Case Analysis in Visual Feature Extraction}.

\subsection{Comparison of Different Settings}
\label{subsec:Comparison of Different Settings}

After Ca-101 is selected as the default detector, Section~\ref{sec:Proposed Approach} provides two contextual encoders, namley the co-occurrence encoder and the layout encoder and two RNN network structures, namely single-directional RNN and bidirectional RNN, with three basic network units, namely simple-RNN unit, GRU and LSTM unit. In the following sections, we will discuss these options and try to find the best combination.

\subsubsection{Performance of Co-occurrence Encoder}
\label{subsub:Performance of Co-occurrence Encoder}

TABLE~\ref{tab:CoEnComp} shows the performance of the co-occurrence encoder combined with different RNN classifiers. For classifiers using simple-RNN and LSTM units, the single-directional structures are better than the bidirectional ones. The reverse is true when GRU is used. Simple-RNN units in both structures clearly outperform the others, which makes it to be the default network unit. The best combination belongs to simple-RNN units with single-directional structure, which is regarded as the best classifier for co-occurrence encoder.

\begin{table}[htbp]
	\centering
	\caption{Performances of co-occurrence encoder in percentage terms.}
	\setlength{\tabcolsep}{12pt}
	\label{tab:CoEnComp}
	\begin{tabular}{cccc}
		\Xhline{1.5pt}
		\noalign{\smallskip}
		\textbf{Combinations} & \textbf{M-P} & \textbf{M-R} & \textbf{M-F1} \\
		\Xhline{1pt}
		\noalign{\smallskip}
		simple-RNN+single-directional & \textbf{81.47} & \textbf{80.53} & \textbf{81.00} \\
		\noalign{\smallskip}
		simple-RNN+bidirectional & 81.13 & 80.20 & 80.66 \\
		\noalign{\smallskip}
		GRU+single-directional & 80.43 & 79.22 & 79.82 \\
		\noalign{\smallskip}
		GRU+bidirectional & 80.57 & 79.24 & 79.90 \\
		\noalign{\smallskip}
		LSTM+single-directional & 80.85 & 79.50 & 80.17 \\
		\noalign{\smallskip}
		LSTM+bidirectional & 80.50 & 79.39 & 79.94 \\
		\noalign{\smallskip}
		\Xhline{1.5pt}
	\end{tabular}
\end{table}

\subsubsection{Performance of Layout Encoder}
\label{subsub:Performance of Layout Encoder}

TABLE~\ref{tab:LayVSCo} shows the comparison between co-occurrence encoder and layout encoder combined with single-directional and bidirectional structure using simple-RNN units. Layout encoder clearly beats co-occurrence encoder, indicating that the spatial arrangement of the building reflects a certain structural context and is useful for distinguishing street view images with different types of land use. For layout encoders, the structure of RNN does not matter much. This also indicates that the spatial arrangement of buildings has a certain robustness for distinguishing different land use scenes.

\begin{table}[htbp]
	\centering
	\caption{Performances of co-occurrence and layout encoder in percentage terms.}
	\setlength{\tabcolsep}{11pt}
	\label{tab:LayVSCo}
	\begin{tabular}{cccc}
		\Xhline{1.5pt}
		\noalign{\smallskip}
		\textbf{Combinations} & \textbf{M-P} & \textbf{M-R} & \textbf{M-F1} \\
		\Xhline{1pt}
		\noalign{\smallskip}
		co-occurrence+single-directional & 81.47 & 80.53 & 81.00 \\
		\noalign{\smallskip}
		co-occurrence+bidirectional & 81.13 & 80.20 & 80.66 \\
		\noalign{\smallskip}
		layout+single-directional & 81.66 & \textbf{81.02} & 81.34 \\
		\noalign{\smallskip}
		layout+bidirectional & \textbf{81.81} & 80.94 & \textbf{81.37} \\
		\noalign{\smallskip}
		\Xhline{1.5pt}
	\end{tabular}
\end{table}

\subsubsection{RNN Training Using Ground Truth Bounding Boxes}
\label{subsub:RNN Training Using Ground Truth Bounding Boxes}

From Fig.~\ref{fig:PipeLine} we know that our RNN is trained by the bounding boxes output from detectors. Why don’t we use the ground truth bounding boxes to train the RNN and use outputs of detectors during test stage? TABLE~\ref{tab:GTTrain} shows the comparison between using and not using the ground truth bounding boxes during training stage. The results are disappointing. The “standard answer” seems to be helpless might due to the mismatch during training and test stage.

\begin{table}[htbp]
	\centering
	\caption{Comparison between using and not using the ground truth bounding boxes for training.}
	\setlength{\tabcolsep}{13pt}
	\label{tab:GTTrain}
	\begin{tabular}{cccc}
		\Xhline{1.5pt}
		\noalign{\smallskip}
		\textbf{Training Combinations} & \textbf{M-P} & \textbf{M-R} & \textbf{M-F1} \\
		\Xhline{1pt}
		\noalign{\smallskip}
		co-occurrence+ground truth & 77.76 & 71.65 & 74.58 \\
		\noalign{\smallskip}
		co-occurrence+Ca101 best & 81.47 & 80.53 & 81.00 \\
		\noalign{\smallskip}
		layout+ground truth & 80.03 & 80.93 & 80.48 \\
		\noalign{\smallskip}
		layout+Ca101 best & \textbf{81.81} & \textbf{80.94} & \textbf{81.37} \\
		\noalign{\smallskip}
		\Xhline{1.5pt}
	\end{tabular}
\end{table}

Comparing the co-occurrence coding and layout coding, it can be found that the influence of training-test mismatch on the latter (0.89\% in M-F1) is significantly lower than that on the former (6.42\% in M-F1). This observation once again demonstrates the robustness of spatial structure. So far, the optimal performance of the proposed approach is generated by layout coding combined with bidirectional RNN structure.

\subsection{Comparison with Baseline}
\label{subsec:Comparison with Baseline}

Finally, the upper limit and the optimal performance of proposed approach and baseline are compared in TABLE~\ref{tab:VSBaseline}. In Section~\ref{subsub:RNN Training Using Ground Truth Bounding Boxes}, the poor performance has been shown when using ground truth bounding boxes as training samples but the outputs of a detector as test samples, due to the training-test mismatch. How about we use the ground truth bounding boxes also in the test stage? It is impossible for a classification system to know some intermediate results of test samples in advance (e.g., ground truth bounding boxes of buildings in the test sample images), but the hypothesis could help us to find out the performance upper limit of the proposed encoder-classifier system. Upper limit means that a perfect detector is used, whose outputs during training and test stages are ground truth bounding boxes which make the proposed encoder-classifier system perform best.

\begin{table}[htbp]
	\centering
	\caption{The upper limit, proposed approach and baseline.}
	\setlength{\tabcolsep}{15pt}
	\label{tab:VSBaseline}
	\begin{tabular}{cccc}
		\Xhline{1.5pt}
		\noalign{\smallskip}
		\textbf{Models} & \textbf{M-P} & \textbf{M-R} & \textbf{M-F1} \\
		\Xhline{1pt}
		\noalign{\smallskip}
		layout+perfect detector & \underline{95.54} & \underline{92.15} & \underline{93.82} \\
		\noalign{\smallskip}
		layout+Ca101 best & \textbf{81.81} & \textbf{80.94} & \textbf{81.37} \\
		\noalign{\smallskip}
		base line: ResNet50 & 69.16	& 68.94	& 69.05 \\
		\noalign{\smallskip}
		\Xhline{1.5pt}
	\end{tabular}
\end{table}

\begin{figure*}[!htb]
	\centering
	\includegraphics[width=0.3\linewidth]{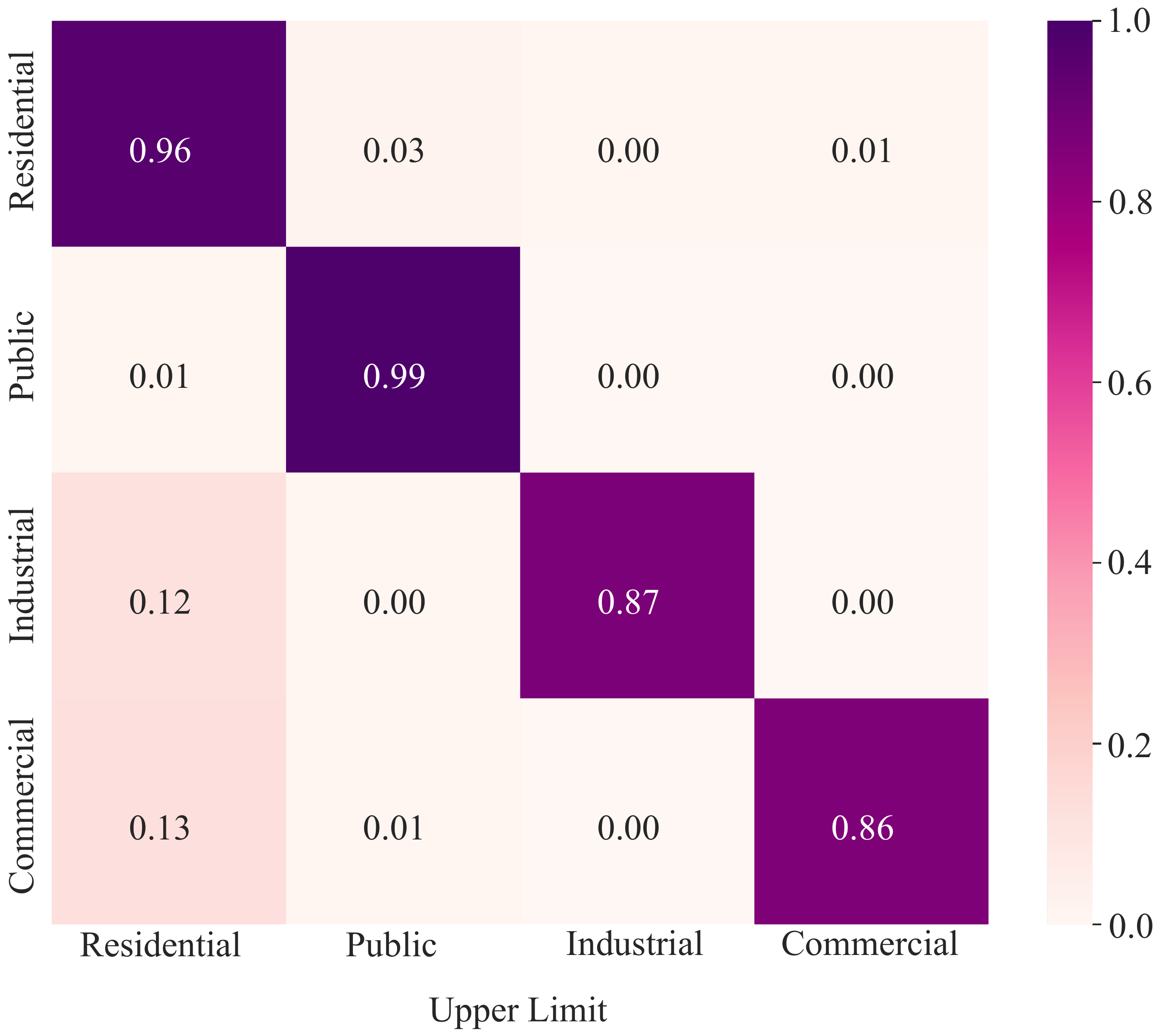}
	~~\includegraphics[width=0.3\linewidth]{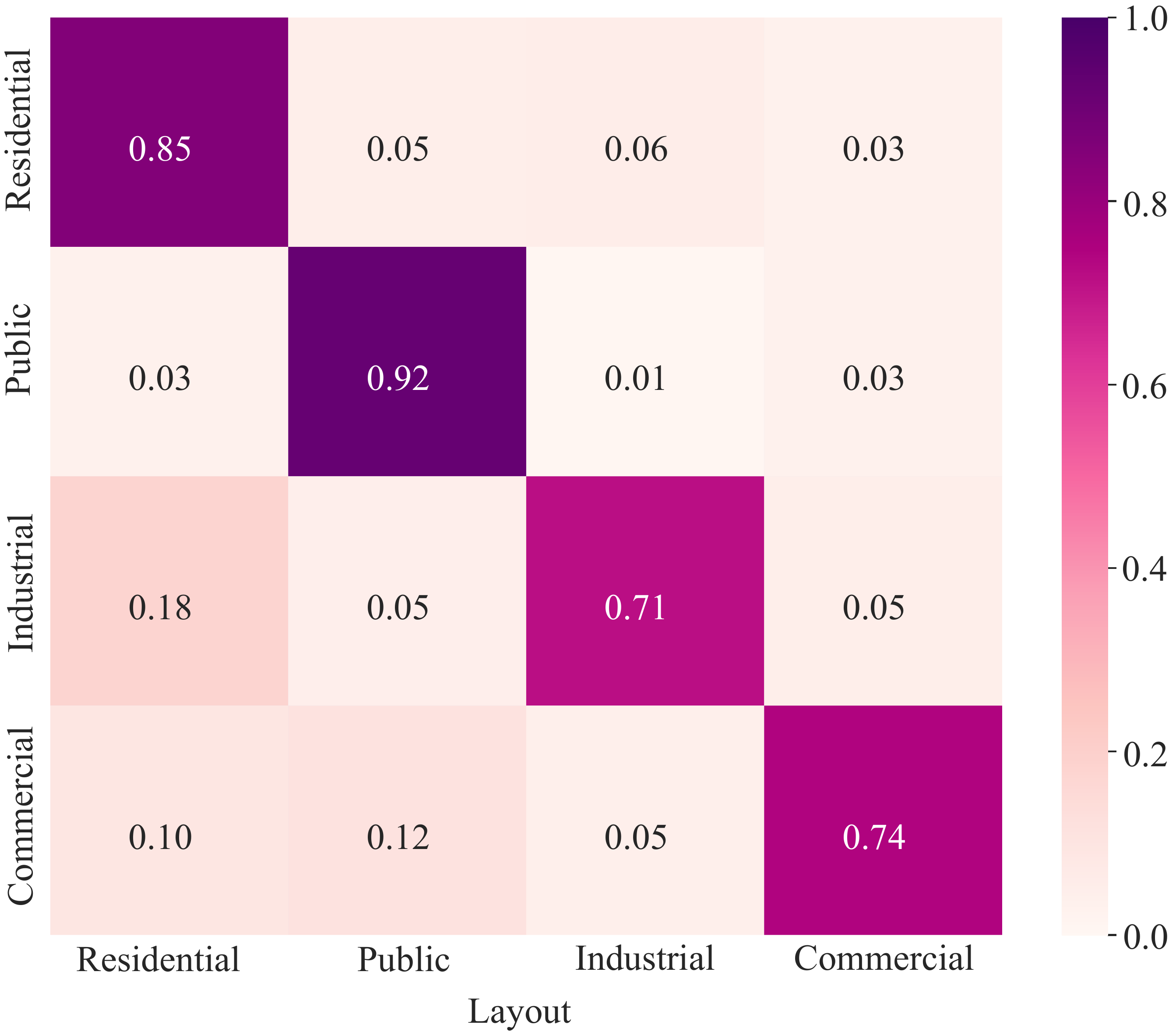}
	~~\includegraphics[width=0.3\linewidth]{images/ResNet50Confusion.pdf}
	\caption{The confusion matrices of upper limit (left), optimal performance of proposed approach (middle) and baseline (right).}
	\label{fig:3Con}
\end{figure*}

The M-F1 of the perfect detector shows a 12.45\% higher than the current optimal combination of the proposed approach, which means that our approach has a lot of room to improve with the progress of object detection. On the other hand, the M-F1 of the proposed approach presents a 12.32\% higher than the baseline (ResNet50 image classification model), which is a significant improvement. More details could be obtained in the confusion matrices of the four classes of land use scenes (Fig.~\ref{fig:3Con}). The category with the most room for improvement of the proposed method is industrial (16\%). While the category with the most improvement over the baseline is public (39\%). These are discussed in more details in the following parts.

\subsubsection{Typical Case Analysis in Visual Feature Extraction}
\label{subsub:Typical Case Analysis in Visual Feature Extraction}

\begin{figure*}[htbp]
	\centering
	\subfigure[A case of \textit{public}. The prediction of ResNet50 is \textit{residential}. The prediction probability of \textit{public} is only 0.011.]
	{\begin{minipage}[htbp]{0.8\textwidth}
			\includegraphics[width=\linewidth]{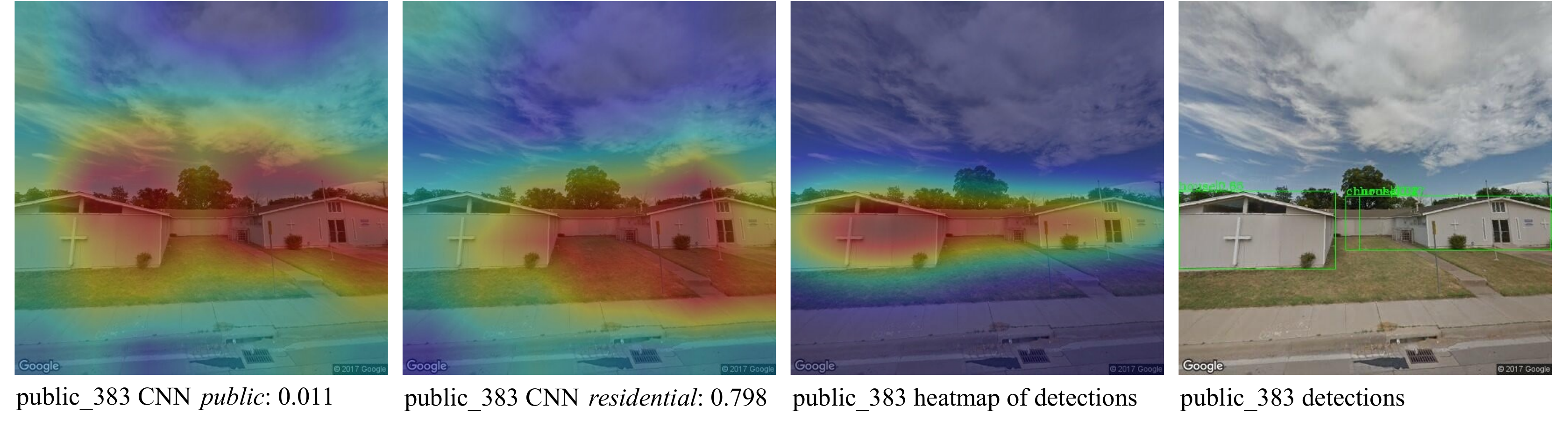} 
			\label{fig:HeatPub}       
		\end{minipage}
	}
	\subfigure[A case of \textit{industrial}. The prediction of ResNet50 is \textit{commercial}. The prediction probability of \textit{industrial} is only 0.142.]
	{\begin{minipage}[htbp]{0.8\textwidth}
			\includegraphics[width=\linewidth]{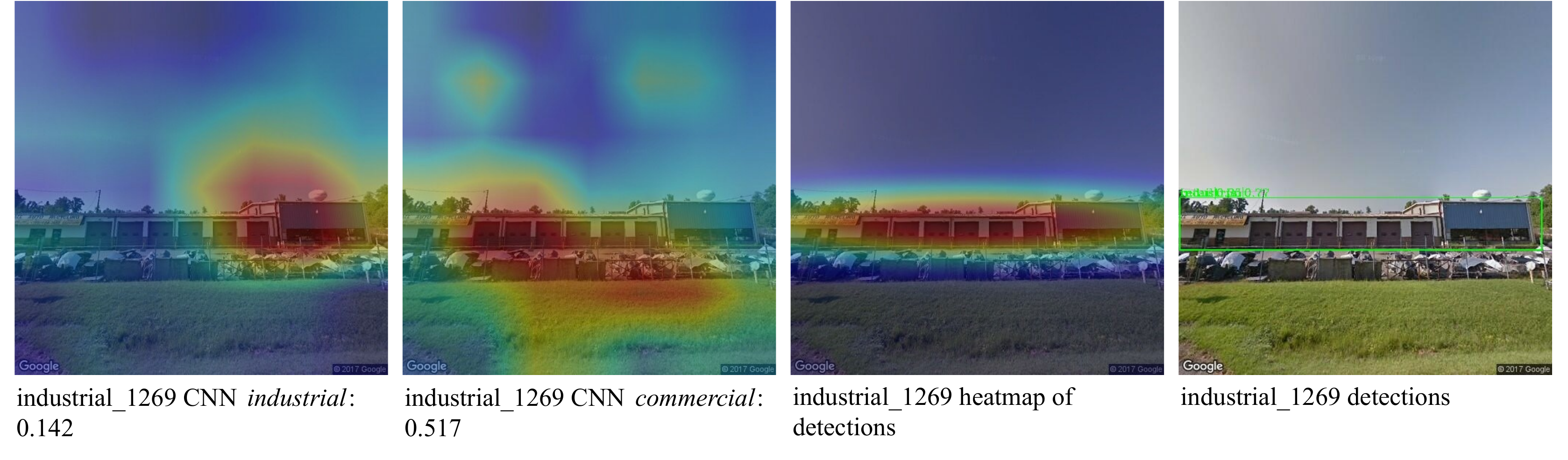} 
			\label{fig:HeatInd}      
		\end{minipage}
	}
	\caption{Heatmaps of the correct class (left 1) and prediction (left 2) by ResNet50 and detections (left 3).}
	\label{fig:Heatmap}
\end{figure*}

In this part, we try to explain why the significant improvement by proposed approach over common image-level end-to-end visual models in terms of the effectiveness of visual feature extraction. In order to demonstrate the effectiveness of visual features more intuitively, we use visual feature heatmap~\cite{selvaraju2017grad}, which is often used for interpretability analysis of neural networks. For baseline (ResNet50), feature maps before the last average pooling layer are used to generate the heatmaps. The regions that contribute to the prediction of each class are marked by warm color regions in the heatmap of the class. For the proposed approach, the approximate heatmaps are generated by the outputs of detector (Fa-50). For each bounding box $b_i$, its approximate heatmap is assumed to be a two-dimensional Gaussian distribution that described by~(\ref{eq:Heatmap}), where $T_{x, y}$ is the temperature of point $(x,y)$ in an image, $w_i$, $h_i$ are the width and height of $bi$ and $(x_{i0}, y_{i0})$ are the center coordinates of $b_i$. Since the detector was not directly used for scene classification, we overlaid and normalized the heatmap of each detection to show the regions that were potentially helpful for the final classification. Heatmaps of typical cases are shown in Fig.~\ref{fig:Heatmap}.

\begin{equation}\label{eq:Heatmap}
	T_{x, y}=\frac{1}{\pi \sqrt{w_{i} h_{i}}} \exp \left\{-2\left[\frac{\left(x-x_{i 0}\right)^{2}}{w_{i}^{2}}+\frac{\left(y-y_{i 0}\right)^{2}}{h_{i}^{2}}\right]\right\}
\end{equation}

We chose typical cases from the categories with the most improvement over the baseline namely public (39\%) and industrial (5\%) to do an in-depth analysis. Fig.~\ref{fig:HeatPub} shows a case of \textit{public} (image ID: public\_383 in BEAUTY). The activated regions of the heatmap for the correct class (\textit{public}) include large areas of sky and ground. While the activated regions of the heatmap for the prediction (\textit{residential}) do not cover the whole buildings and miss the cross of the church in left, which carries key information about land use. It also contains lots of areas of ground. In contrast, the heat map generated by detector covers all key regions tightly. A similar situation can be clearly demonstrated in the \textit{industrial} example (Fig.~\ref{fig:HeatInd}).

\subsubsection{Typical Case Analysis in Context Information Extraction}
\label{subsub:Typical Case Analysis in Context Information Extraction}

\begin{figure*}[htbp]
	\centering
	\subfigure[A case of \textit{commercial}. Wrong prediction happened after three bounding boxes are tampered with.]
	{\begin{minipage}[htbp]{0.8\textwidth}
			\includegraphics[width=\linewidth]{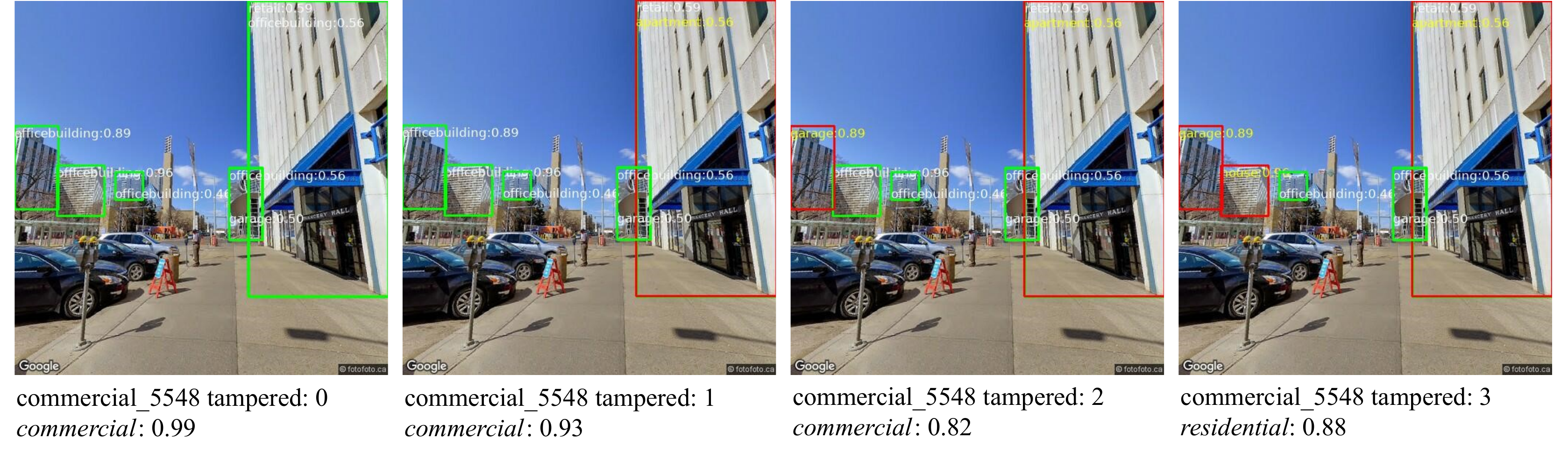} 
			\label{fig:TamCom}       
		\end{minipage}
	}
	\subfigure[A case of \textit{residential}. Wrong prediction happened after four bounding boxes are tampered with.]
	{\begin{minipage}[htbp]{0.8\textwidth}
			\includegraphics[width=\linewidth]{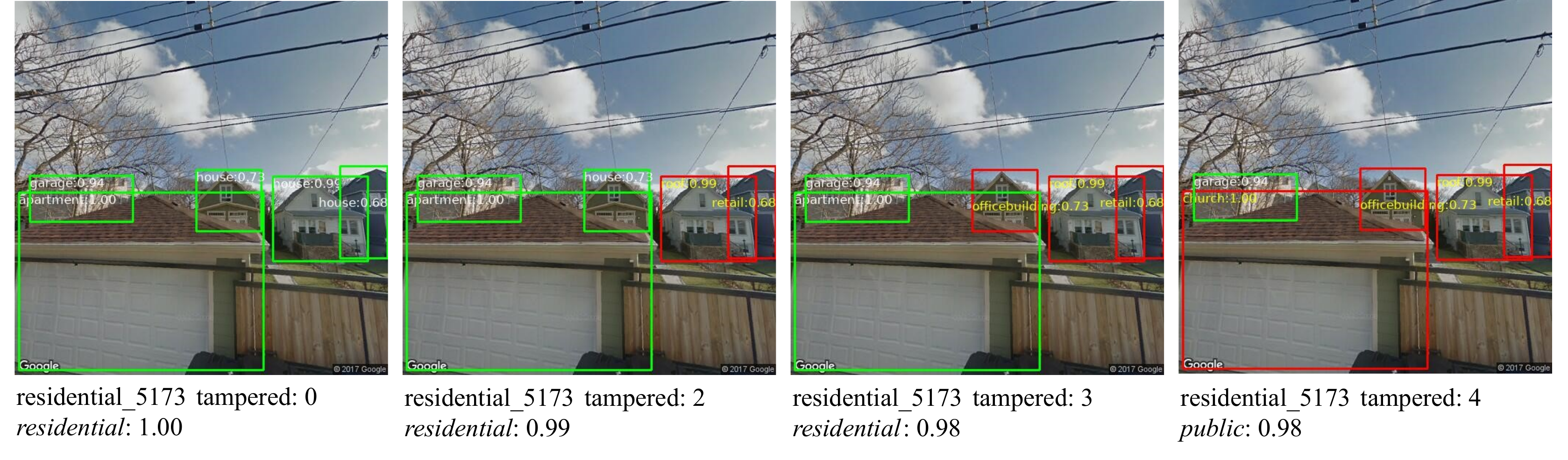} 
			\label{fig:TamRes}      
		\end{minipage}
	}
	\caption{The prediction of proposed approach keeps correct when the detector makes a small number of wrong predictions (red bounding boxes).}
	\label{fig:Tampered}
\end{figure*}

Cases in previous part demonstrate that the proposed approach can obtain more effective visual features than CNN-based image-level end-to-end model by using detectors specially trained for buildings. Is the good performance of the proposed approach entirely dependent on the detector? In this part, several cases will show that the proposed encoder and RNN-based classifier can obtain correct scene classification according to the learned context information, even if the detector incorrectly predicts the class of some buildings. Outputs of the detector are firstly encoded by the proposed “CODING”. Then a growing number of bounding boxes with high confidence scores have been tampered with as the ones of wrong categories. During this process, the results of the final scene classification are observed all the time. Cases are shown in Fig.~\ref{fig:Tampered}.

For illustration purposes, only bounding boxes with a confidence score greater than 0.4 are shown. The actual number of detections of the first case “commercial\_5548” is 12, showing only 7 in Fig.~\ref{fig:TamCom}. Parts of the bounding boxes overlap, such as “\textit{retail}: 0.95” and “\textit{office building}: 0.56” on the far right, and “\textit{office building}: 0.56” and “\textit{garage}: 0.50” next to them. Several bounding boxes are tampered with to the wrong class one by one. When two bounding boxes were tampered with (\textit{office building}: 0.56 $\rightarrow$ \textit{apartment}: 0.56, \textit{office building}: 0.89 $\rightarrow$ \textit{garage}: 0.89), the proposed approach could still make correct prediction about the scene class. When the third bounding box was tampered with (\textit{office building}: 0.96 $\rightarrow$ \textit{house}: 0.96), the prediction jumped from \textit{commercial}: 0.82 to \textit{residential}: 0.88 without a gradient.
In the second case “residential\_5173”, there are some errors in the original detections. The ground truth of the largest bounding box in the lower left corner is \textit{garage}, but the detection is \textit{apartment}: 1.00. The ground truth of the small bounding box on the left is \textit{apartment}, but the detection is \textit{garage}: 0.94. This kind of detection errors could cause minor changes to the layout encoding and will be ignored in RNN-based classifier. And since the co-occurrence relationship between building classes does not change, it did not affect the final classification result (\textit{residential}: 1.00). In Fig.~\ref{fig:TamRes}, class of the detections are tampered with one by one from right to left. The prediction jumped from \textit{residential}: 0.98 to \textit{public}: 0.98 after four bounding boxes were tampered with. Some bounding boxes that were not drawn because their confidence scores were less than 0.4 still contributed to the context relationships such as co-occurrence and layout, which resulted in the prediction of scene class being maintained when three detections were tampered with. The last straw was the manipulation of the bounding box with the largest size and confidence score (\textit{apartment}: 1.00 $\rightarrow$ \textit{church}: 1.00).

The case analysis above can give a glimpse of why the proposed approach achieved better performance over image-level end-to-end CNN models such as ResNet50. The general conclusions are given in Section~\ref{sec:Conclusion and Future Work}.

\subsection{Use on Open World Street View Image Data}
\label{subsec:Land Use Maps Generation}

\begin{figure*}[htbp]
	\centering
	\includegraphics[width=0.75\linewidth]{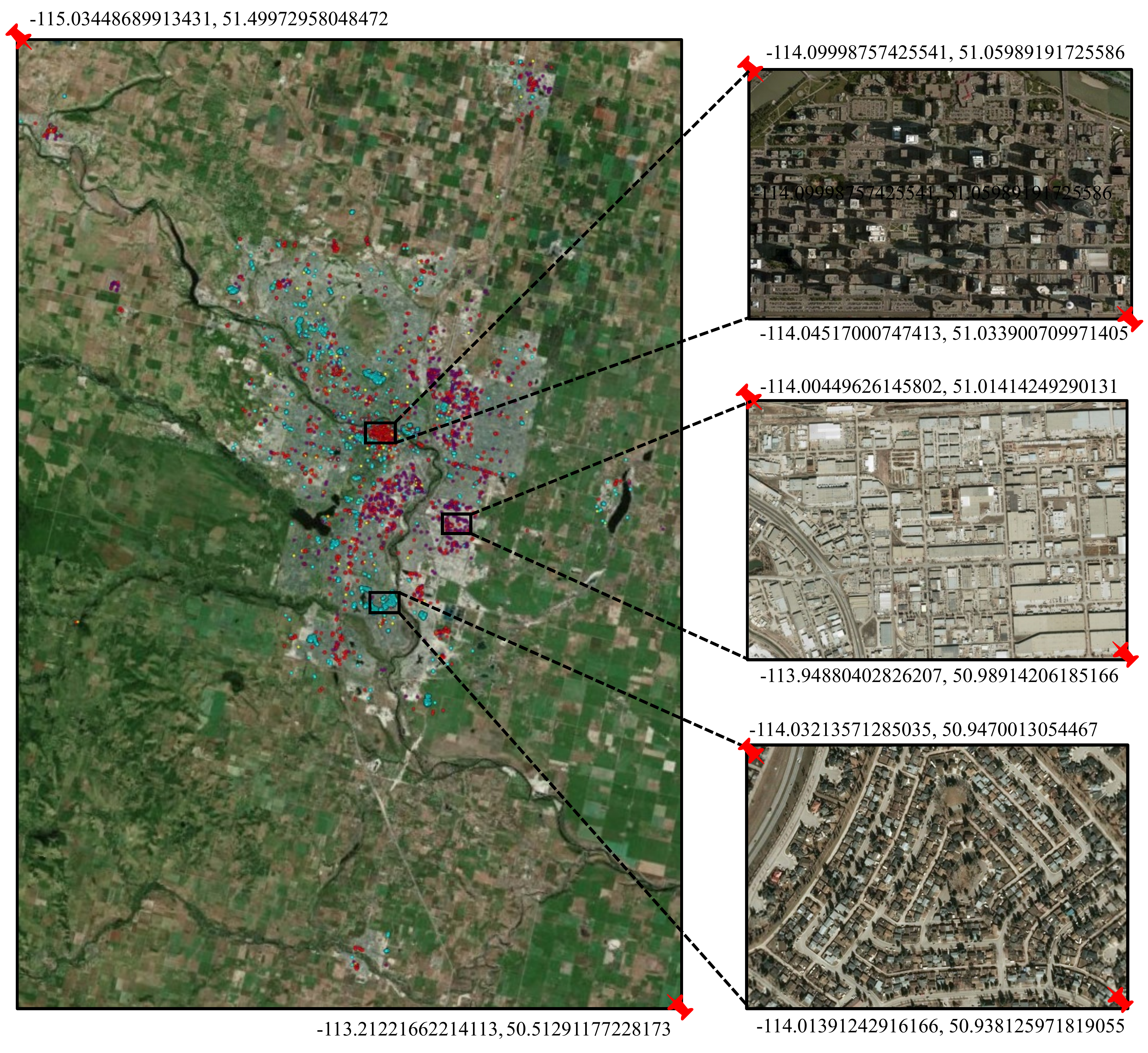} 
	\caption{The city-scale land use classification map of Calgary.}
	\label{fig:Calgary}
\end{figure*}

To further verify the performance of the proposed approach on an open world data set, land use maps of Calgary are generated using open world GSV images provided by~\cite{kang2018building}. Land use maps of Calgary based on 6,124 street view images are shown in Fig.~\ref{fig:Calgary}. Geo-tagged street view images were classified by the proposed approach. The results were then drawn on CesiumJS\footnote{\url{https://cesium.com/cesiumjs/}} according to the geographical locations of the input images. Four land use classes \textit{residential}, \textit{commercial}, \textit{industrial} and \textit{public} are marked by dots of blue, red, purple and yellow respectively. Regions where dots with the same color clustered in a city-scale map are zoomed in to see if the classification is correct.

Generally speaking, the distribution of \textit{residential} area, \textit{commercial} area and \textit{industrial} area in Calgary is relatively balanced. The \textit{commercial} areas (red dots) are relatively concentrated, while the other two are scattered. In Fig.~\ref{fig:Calgary}, a \textit{commercial} area, an \textit{industrial} area and a \textit{residential} area that was shown as red, purple and blue dots cluster respectively were zoomed in. As can be seen from the zoomed over-head image on the top right, the buildings of \textit{commercial} area are dominated by tall buildings. In the middle right zoomed image, the \textit{industrial} area is dominated by large flat-roofed buildings with low floors, which is a typical feature of the \textit{industrial} area. In zoomed over-head image on the bottom right, the characteristics of \textit{residential} areas are also obvious for a large number of small well-arranged low-rise buildings. The \textit{public} areas represented by yellow dots rarely form clusters.

\section{Conclusion and Future Work}
\label{sec:Conclusion and Future Work}

%\subsection{Conclusion}
%\label{subsec:Conclusion}

As the CNNs gradually show an overwhelming advantage in common visual tasks, various image-level CNN models are increasingly favored in street view image classification in recent years. In this paper, a dataset “BEAUTY” is presented, which can be used for both street view image classification and building detection. We used ResNet50, which performs steadily and well on common visual classification tasks as a baseline model to represent the current mainstream image-level CNN models. However, the macro-precision of the ResNet50 was only 69.16\%. After analysing large number of street view image samples, we find that the approaches based on image-level CNN models have the following fatal problems.

\begin{itemize}
	\item The undifferentiated use of the whole image leads to the extraction of common visual factors that confuse classification.
	
	\item Street view image labels for land use classification are often concepts with a high level of abstraction and cannot be described directly and effectively with visual features.
\end{itemize}

As can be seen from the example in Fig.~\ref{fig:Heatmap}, although CNNs have the ability to extract regions conducive to classification through autonomous learning, these regions are often not accurate when classification labels cannot be directly and effectively described with visual features. In addition, the only use of visual semantics (e.g., the recognition results of objects) can no longer well represent highly abstract land-use concepts, which must be done with context-describing visual syntax. Based on the above considerations, this paper proposes a “Detector-Encoder-Classifier” architecture. Object detectors extract visual features that are more recognizable by learning the annotations specifically for buildings. The proposed “CODING” method encodes the context relations such as co-occurrence and layout of these highly recognizable visual objects. At last, RNNs are very suitable for accurately classifying the combination patterns of visual elements with structural relations. The proposed approach performs 81.81\% on macro-precision, an improvement of 12.56\% over the baseline model.

%\subsection{Future Work}
%\label{subsec:Future Work}

The first row of TABLE~\ref{tab:VSBaseline} shows the performance of the proposed approach using a “perfect detector”, which gives the upper limit of the proposed approach and two ideas for improving the performance under the current architecture.

\begin{itemize}
	\item To achieve the upper limit, better detectors are needed. With the development of object detection, this plug-and play-module can be upgraded continuously.
	
	\item To exceed this limit, more powerful context encoders need to be proposed. Self-attention~\cite{cheng2016long} or transformers model~\cite{vaswani2017attention} might be used.
\end{itemize}

In addition to models, data form different sources are also an important way to improve the performance. A more accurate description of the land use may be obtained by matching the layout of the building in street view images to one in overhead images.

\section*{Acknowledgment}
The authors would like to thank the authors of reference~\cite{kang2018building} for publishing the BIC\_GSV dataset including city scale GSV images. We would also like to thank Hongbin Liu and Zhiwei He, the experts in architecture and urban planning from the BIM Research Center, Qingdao Research Institute of Urban and Rural Construction for their professional guidance on manual annotation. Thanks to those who participated in manual annotation for building detection: Yu Ma, Shanshan Lin, Ying Guo and Kaixin Li, and who participated in manual annotation for street view image classification: Ying Zhang, Jiaojie Wang, Shujing Ma and Yue Wang. This work has been supported by the National Natural Science Foundation of China (Grant No. 61701272).

\ifCLASSOPTIONcaptionsoff
  \newpage
\fi

\bibliographystyle{IEEEtran}
\bibliography{bibfile}

\end{document}